\documentclass[pdflatex,sn-mathphys,referee]{sn-jnl}

\newcommand{\etal}{\textit{et al. }}

\usepackage{subfigure}
\usepackage{multicol}
\usepackage{multirow}

\usepackage{color}
\definecolor{red}{rgb}{1.00,0.00,0.00}
\definecolor{blue}{rgb}{0.00,0.00,1.00}
\newcommand{\cred}[1] {\textcolor{red}{#1}}
\newcommand{\cblue}[1] {\textcolor{blue}{#1}}

\jyear{2021}%

\theoremstyle{thmstyleone}%
\theoremstyle{thmstyletwo}%

\theoremstyle{thmstylethree}%

\begin{document}

\title[Compact Twice Fusion Network for Edge Detection]{Compact Twice Fusion Network for Edge Detection}

\author[1]{\fnm{Yachuan} \sur{Li}}\email{liyachuan@s.upc.edu.cn}

\author[1]{\fnm{Zongmin} \sur{Li}}\email{lizongmin@upc.edu.cn}

\author[2,5]{\fnm{Xavier} \sur{Soria P.}}\email{xavier.soria@unach.edu.ec}

\author[1]{\fnm{Chaozhi} \sur{Yang}}\email{yang.chaozhi@foxmail.com}

\author[1]{\fnm{Qian} \sur{Xiao}}\email{xiaoqian@s.upc.edu.cn}

\author[1]{\fnm{Yun} \sur{Bai}}\email{B22070025@s.upc.edu.cn}

\author[3]{\fnm{Hua} \sur{Li}}\email{lihua@ict.ac.cn}

\author*[4]{\fnm{Xiangdong} \sur{Wang}}\email{wangxiangdong@jmu.edu.cn}

\affil[1] {\orgdiv{College of Computer Science and Technology},\orgname{China University of Petroleum (East China)}, \orgaddress{\street{Changjiang West Road},\city{Qingdao}, \postcode{266500}, \state{Shandong}, \country{China}}}

\affil[2] {\orgdiv{Faculty of Educational Science, Humanities, and Technology},\orgname{National University of Chimborazo}, \orgaddress{\street{Av. Eloy Alfaro}, \city{Riobamba}, \postcode{060110}, \state{Chimborazo}, \country{Ecuador}}}

\affil[3]{\orgname{Institute of Computing Technology Chinese Academy of Sciences}, \orgaddress{\street{South Road Zhongguancun}, \city{Beijing}, \postcode{100190}, \country{China}}}

\affil[4]{\orgdiv{Physical Education Institute}, \orgname{Jimei University}, \orgaddress{\street{Yinjiang Rd}, \city{Xiamen}, \postcode{361021}, \state{Fujian}, \country{China}}}

\affil[5] {\orgdiv{CIDIS},\orgname{ESPOL Polytechnic University}, \orgaddress{\street{Campus Gustavo Galindo}, \city{Guayaquil}, \postcode{090112}, \state{Guayas}, \country{Ecuador}}}


\abstract{The significance of multi-scale features has been gradually recognized by the edge detection community. However, the fusion of multi-scale features increases the complexity of the model, which is not friendly to practical application. In this work, we propose a Compact Twice Fusion Network (CTFN) to fully integrate multi-scale features while maintaining the compactness of the model. CTFN includes two lightweight multi-scale feature fusion modules: a Semantic Enhancement Module (SEM) that can utilize the semantic information contained in coarse-scale features to guide the learning of fine-scale features, and a Pseudo Pixel-level Weighting (PPW) module that aggregate the complementary merits of multi-scale features by assigning weights to all features.
Notwithstanding all this, the interference of texture noise makes the correct classification of some pixels still a challenge. For these hard samples, we propose a novel loss function, coined Dynamic Focal Loss, which reshapes the standard cross-entropy loss and dynamically adjusts the weights to correct the distribution of hard samples. We evaluate our method on three datasets, i.e., BSDS500, NYUDv2, and BIPEDv2. Compared with state-of-the-art methods, CTFN achieves competitive accuracy with less parameters and computational cost. 
Apart from the backbone, CTFN requires only 0.1M additional parameters, which reduces its computation cost to just 60\% of other state-of-the-art methods.
The codes are available at \href{https://github.com/Li-yachuan/CTFN-pytorch-master}{https://github.com/Li-yachuan/CTFN-pytorch-master}.}




\maketitle

\section{Introduction}\label{sec1}

The purpose of edge detection is to extract object boundaries and salient edges from natural images to preserve the key information and ignore insignificant details. Therefore, it is considered a fundamental task in computer vision and plays an important role in higher-level tasks such as salient detection~\cite{zhao2019egnet}, semantic segmentation~\cite{liu2021dance}, and depth map prediction~\cite{zhu2020edge}.

Edge detection which divides the pixels into edge and non-edge is a sub-task of semantic segmentation. Therefore, the pixels classification is the essential question in edge detection. The pioneering works finish this work by local features, such as brightness, gradient and color. The lack of global information limits the performance of edge detection methods, until the advent of Holistically-nested Edge Detection(HED)~\cite{xie2017holistically}. As the pioneer of contemporary edge detection, HED first introduces the deep supervision mechanism to edge detection and learns multi-scale predictions holistically. On this basis, a series of excellent methods are produced~\cite{liu2019richer,he2022bi-directional,deng2018learning,gao2021bottom}.
\begin{figure}[ht]
	\centering
	
	\subfigure[]{
		\centering
		\includegraphics[width=3.6cm,height=2.4cm]{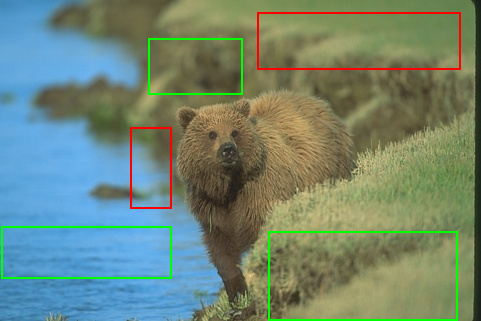}
	}
	\subfigure[]{
		\centering
		\includegraphics[width=3.6cm,height=2.4cm]{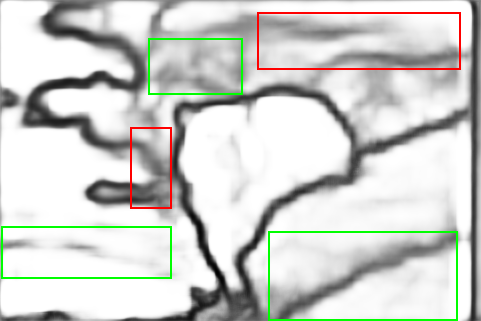}
	}
	\subfigure[]{
		\centering
		\includegraphics[width=3.6cm,height=2.4cm]{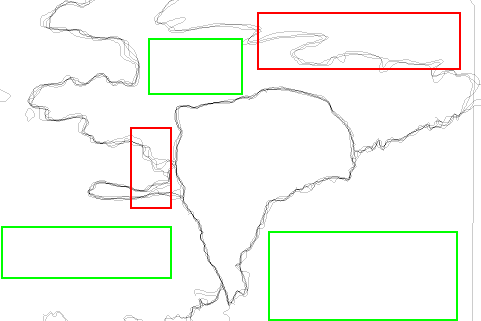}
	}
    \label{hardsample}
	\caption{Illustration of the hard samples. The locations of the hard samples are marked with boxes. (a) an image from the BSDS500 dataset, (b) The result predicted by our method, (c) the corresponding ground truth.}
\end{figure}
Although the performance of edge detection has been significantly improved, these methods suffer from two major issues. Firstly, the number of model parameters increases sharply. With the further exploration of multi-scale features, the model parameters are also greatly increased, which is unable to meet the demands of downstream tasks. In most situations, spending a mass of space and computing resources for a little accuracy improvement doesn't make sense. Secondly, lack of attention to hard samples. Hard samples refer to the pixels whose classification probability is significantly different from the ground truth, that is, the pixels prone to misclassification. As shown in Fig.~\ref{hardsample}, the confidence of the edges in the red box decreases due to texture interference, while the textures in the green box are mistaken for edges. These are two typical examples of hard samples. Hard samples determine the ceiling of detection accuracy, so extra attention should be attached to them.

Aiming to fully exploit multi-scale feature fusion and avoid the sharply increased parameters, we introduce a Compact Twice Fusion Network (CTFN) for edge detection, in which higher quality edges are obtained by fusing the multi-scale features twice. A lightweight Semantic Enhancement Module (SEM) is introduced in the first feature fusion. In SEM, high-level semantic information is used to increase the receptive field of fine-scale features, thereby improving the discrimination of fine-scale branches. However, SEM is a cascade structure based on FPN~\cite{lin2017feature}, in which the high-level semantic information is gradually attenuated in the process of transmission, so the second feature fusion is required to aggregate all scale information. In the second feature fusion, we introduce a Pseudo Pixel-level Weighting (PPW) module, which sets the weights of multi-scale features according to their context information and further reduces the module complexity by decomposing the weights into the spatial weights and the channel weights.

To further enhance the attention for the hard samples, we propose Dynamic Focal Loss (DFL). DFL reshapes the standard cross-entropy loss and dynamically adjusts the weight of the loss assigned to hard samples. Increasing the weights of hard samples is an effective method to optimize hard samples, but effectively determining the hard samples become a new problem. Focal loss~\cite{lin2017focal} discriminates hard samples by the gap between model output and ground truth. However, due to the existence of randomly initialized modules, the output of the model is chaotic at the initial stage of training, so it is unreasonable to identify hard samples by early output. Therefore, DFL distrusts the model output at first and dynamically increases the confidence margin to reduce the adverse effects caused by the early chaos of the model.

The main contributions of the paper are summarized below:
\begin{itemize}
\item We systematically analyze the existing deep-learning based edge detection methods and find two urgent problems.
\item We propose a Compact Twice Fusion Network (CTFN) that fully fuse multi-scale features while maintaining model compactness. CTFN utilizes only 0.1M additional parameters beyond the backbone, resulting in a 60\% reduction in computational cost compared to state-of-the-art methods.
\item For the hard samples in edge detection, we introduce Focal Loss for the first time and propose a Dynamic Focal Loss to solve the chaotic output problem in the early training stage of Focal Loss.
\item Extensive experiments are conducted on BSDS500, NYUDv2, and BIPEDv2 datasets to demonstrate the effectiveness and robustness of our method.
\end{itemize}

\section{Related work}\label{sec2}

The origin of edge detection can be traced back to the last century. Early pioneering methods~\cite{bertasius2015deepedge,kokkinos2015pushing,shen2015deepcontour} mainly focus on local cues, which prevents methods from distinguishing between texture and edge.
In recent years, deep learning based methods come from behind. Edge detection has entered the era of deep learning. We review the recent development of deep learning-based edge detection methods in terms of model structure and loss function.

\subsection{Model structure}

\begin{figure}[htbp]
	\centering
	
	\subfigure[]{
		\centering
		\includegraphics[width=2.5cm,height=2.5cm]{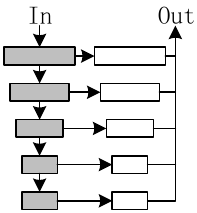}
		\label{stru-cmp1}
	}
	\subfigure[]{
		\centering
		\includegraphics[width=2.cm,height=2.5cm]{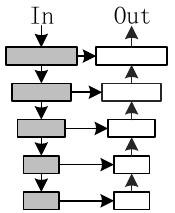}
		\label{stru-cmp2}
	}
	\subfigure[]{
		\centering
		\includegraphics[width=2.5cm,height=2.5cm]{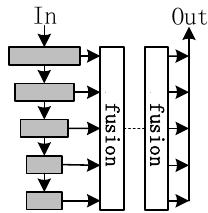}
		\label{stru-cmp3}
	}
	\caption{Comparison of common structures in edge detection. (a) HED-based method, (b) UNet-based method, (c) Multiple Feature Fusion (MFF) method.}
\end{figure}

Holistically-nested Edge Detection (HED)~\cite{xie2017holistically} introduces the deep supervision mechanism to edge detection and learns multi-scale predictions holistically. Inspired by the great success of HED, the most recent methods~\cite{liu2019richer,poma2020dense,huan2021unmixing} are committed to end-to-end learning and enrichment of global features. Multi-scale information in this kind of method is almost independent and final edge maps are obtained only by taking the weighted sum of multi-scale information, as shown in Fig.~\ref{stru-cmp1}. Such methods can be called HED-based methods. It is worth mentioning that although BDCN~\cite{he2022bi-directional} appears to have twice feature fusions, the gradient of the first feature fusion is truncated, so we still regard it as a HED-based method. The success of HED-based methods is undoubted, but the lack of semantic information in fine-scale features due to multi-scale features independence is an open problem worthy of study.

For the aforementioned problem, Wang \etal~\cite{wang2018deep} introduces a new fusion method, in which semantic information from coarse-scale is utilized to facilitate fine-scale feature learning. This kind of method has been further developed in recent years~\cite{deng2018learning,deng2020deep}. Since the structure of these methods is similar to U-Net~\cite{ronneberger2015u}, they can be termed UNet-based methods, as mentioned in Fig.~\ref{stru-cmp2}. UNet-based methods pay more attention to fine-scale features, while the coarse-scale semantic information gradually decay in the process of feature fusion. 

A natural idea is to merge the two structures to fuse the features further. These merged structures are what we refer to as Multiple Feature Fusion (MFF) methods, as illustrated in Fig.~\ref{stru-cmp3}. FCL~\cite{xuan2022fcl} fuses multi-scale features twice. In the first fusion, LSTM is introduced to address attenuation problems during the downward fusion of high-level semantic information. In the second fusion, FCL designs a pixel-level weighting module to assign weights to each feature to refine the feature fusion process. BAN~\cite{gao2021bottom} repeatedly fuses multi-scale features in both bottom-up and self-downward directions to achieve fully fused multi-scale features.
MFF methods significantly improve edge detection accuracy. However, feature fusion dramatically increases model complexity, making them less ideal for downstream tasks and real-time computing.

Inspired by the multiple feature fusion methods~\cite{xuan2022fcl,gao2021bottom}, we propose CTFN, which retains the structure of multiple feature fusion while removing unnecessary high-cost modules.
CTFN not only ensures the full fusion of multi-scale features but also reduces the number of parameters and the computational cost associated with multiple feature fusion.

\subsection{Loss function}

 Weighted Cross-Entropy is employed to supervise the learning of the network in HED~\cite{xie2017holistically}. RCF~\cite{liu2019richer} filters out samples with disputed ground truth based on HED and leads to the most popular loss function.

Since the relatively scarcity of edge pixels compared to non-edge pixels, backpropagation gradient stabilization requires significantly larger weights for edge pixels.
 
And it causes the problem of blurry edges. That is, there is a wide transition region between edge and non-edge, as shown in Fig.~\ref{hardsample}(b). Therefore, the following works are expected to obtain crisp edges by optimizing the loss function.
Deng \etal\cite{deng2018learning} believe Weighted Cross-Entropy prevents generating crisp edges and replace it with the dice coefficient and cross-entropy. In another work~\cite{deng2020deep}, they further consider the structural differences between the output and the ground truth by the structural similarity index (SSIM)~\cite{zhou2004Image}. Huan \etal\cite{huan2021unmixing} directly divided the image into three parts: edge pixels, confusing pixels, and non-edge pixels, and optimize them respectively. 

These loss functions give a further boost to edge detection, but they all ignore the problem of hard samples mentioned above. To this end, we propose a novel Dynamic Focal Loss.

\section{Method}\label{sec:method}

Our innovation can be divided into two parts: network architecture and loss function. In this section, we describe each in detail.

\subsection{Compact Twice Fusion Network}
\begin{figure*}[ht]
    \centering
	\includegraphics[width=0.9\textwidth]{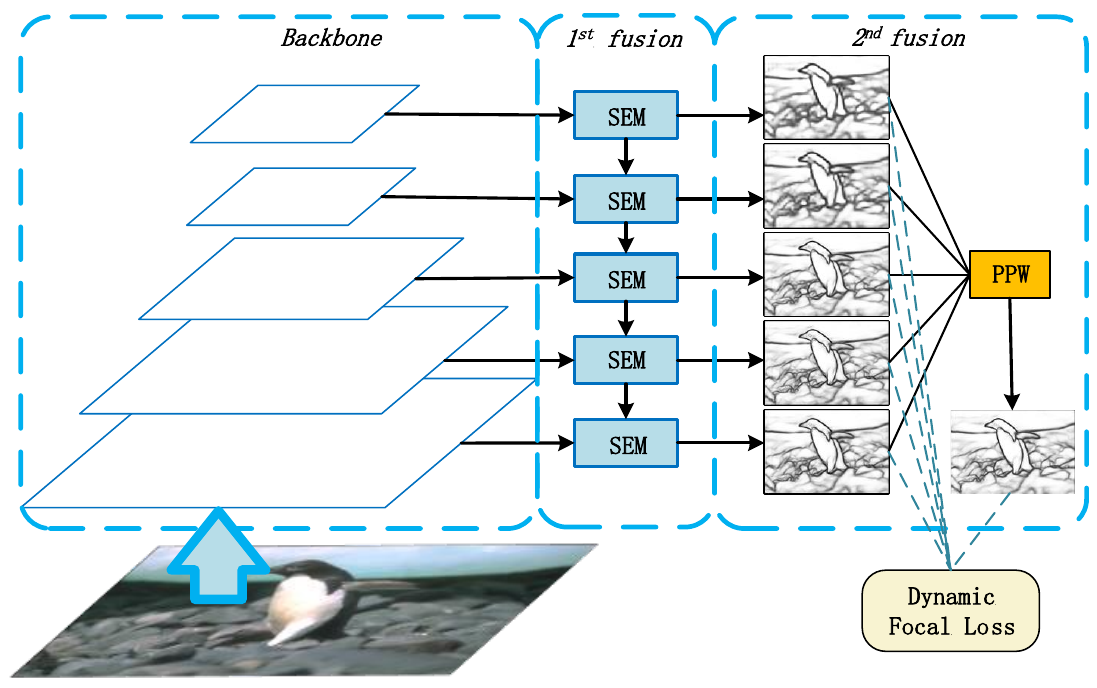}
	\caption{Network architecture of CTFN. The input is an image with arbitrary size, and the output is an edge map with the same size as the input. The first feature fusion stage is donated by $1^{st} fusion$ and $2^{nd} fusion$ stands for the second feature fusion stage.} \label{fig-stru}
\end{figure*}

The overall architecture of our proposed CTFN is shown in Fig.~\ref{fig-stru}, which contains three main stages: backbone, first feature fusion, and second feature fusion. Multi-scale features are obtained through the backbone, and then edges are generated through twice feature fusion.

\subsubsection{Backbone}

 To be fair with existing methods~\cite{xie2017holistically,liu2019richer,he2022bi-directional}, the backbone is based on the VGG16~\cite{simonyan2014very} as well. After removing the last pooling layer and fully connected layers, the 13 convolutional layers in VGG16 are divided into 5 blocks by pooling layers. The dilations in the 5\emph{th} block are set with 2 to enlarge the receptive fields. Backbone is used to generate multi-scale features, which is a prerequisite for two feature fusions and is also the module with the most parameters.
 
\subsubsection{First Feature Fusion} 

In edge detection, identifying the differences between edges and textures is critical but can be challenging due to their similar appearances. Often, the differentiating factor is semantic information, which is closely tied to the receptive field of the feature. The larger the receptive field, the more semantic information that can be captured, thereby enhancing the preservation of edges and suppression of textures. Consequently, increasing the feature receptive field is key to improving feature quality, particularly for fine-scale branches that have smaller receptive fields than other scales.
To address this issue, we introduce a Semantic Enhancement Module (SEM) into the first feature fusion stage. The SEM specifically targets the fine branch features and works to increase their receptive fields, which improves their judgment ability. By enhancing the receptive field of the fine branch features, we can better capture relevant semantic information while minimizing the risk of introducing noise. This strategy allows us to achieve higher accuracy in edge detection tasks and produce sharper, more visually appealing results. 

\begin{algorithm}
\caption{The process of the first feature fusion}
\label{algo-fff}
\begin{algorithmic}[1]
\Require $x_i\in (C,H_i,W_i), i\in [1,2,3,4,5]$
\Ensure $x'_i\in (21,H_i,W_i), i\in [1,2,3,4,5]$
\State $i = 5$
\While{$i > 0$}
    \State $x_i = Conv_{1\times1}(x_i)\in[21,H_i,W_i]$
    \State $x'_i = GroupNorm(x_i)$
    \If{$i\neq 5$}
        \State $x'_{i+1}=Interpolate(x_{i+1})\in [21,H_i,W_i]$
        \State $x'_i=\frac{(x'_{i+1}+x'_i)}{2}$
    \EndIf
    \State i=i-1
\EndWhile
\end{algorithmic}
\end{algorithm}

The detailed process for the first feature fusion is described in Algorithm~\ref{algo-fff}
In the first step, we aim to balance the multi-scale feature channels and reduce computational costs. To achieve this, we use convolution to decrease the number of channels from the original value to 21 while also maintaining a consistent representation of features across scales. In particular, we set the kernel size of convolution to 1 to effectively reduce parameters and ensure computational efficiency. This strategy has been adopted by other works as well, such as~\cite{liu2019richer,su2021pixel}.
After reducing the number of channels, we use the Group Normalization (GN) Layer~\cite{10.1007/978-3-030-01261-8_1} to normalize the features. This step is important because it prevents vanishing gradients during subsequent feature fusion, which can substantially hinder network training and result in poor convergence.
Finally, the normalized features are combined through matrix addition from top to bottom. This approach leverages the benefits of the different features at multiple scale levels and results in a more robust representation of the input data. By performing the feature fusion in this way, we can increase the network's accuracy and reduce noise and other undesirable artifacts that may compromise the integrity of the outputs. Overall, this series of steps culminates in a powerful deep learning framework that can effectively tackle a wide range of edge detection tasks.

In fact, the effectiveness of using coarse-scale branches to guide the learning of fine-scale branches has been verified by previous works~\cite{xuan2022fcl,deng2018learning,gao2021bottom}, but the previous works introduce redundant parameters and calculations in the process of feature fusion, while our proposed SEM only retains the structure of this feature fusion. SEM is an extremely lightweight module that requires minimal additional learnable parameters, typically only a few convolution operations with a kernel size of 1. As a result of this efficient design, SEM not only provides a significant boost to model performance but also results in a compact architecture.

\subsubsection{Second Feature Fusion}

In earlier works~\cite{xie2017holistically,liu2019richer}, the weighted sum of multi-scale edges is directly used to compute the final edge. It is inaccurate that all pixels in the same channel share the same weight and have equal importance in fusion. Recent works~\cite{huan2021unmixing,xuan2022fcl} found this problem and have attempted to assign different weight to each pixel. However, additional operations bring the problem of increasing parameters and decreasing inference speed.

\begin{figure}[ht]
	\centering
	\includegraphics[width=0.8\linewidth]{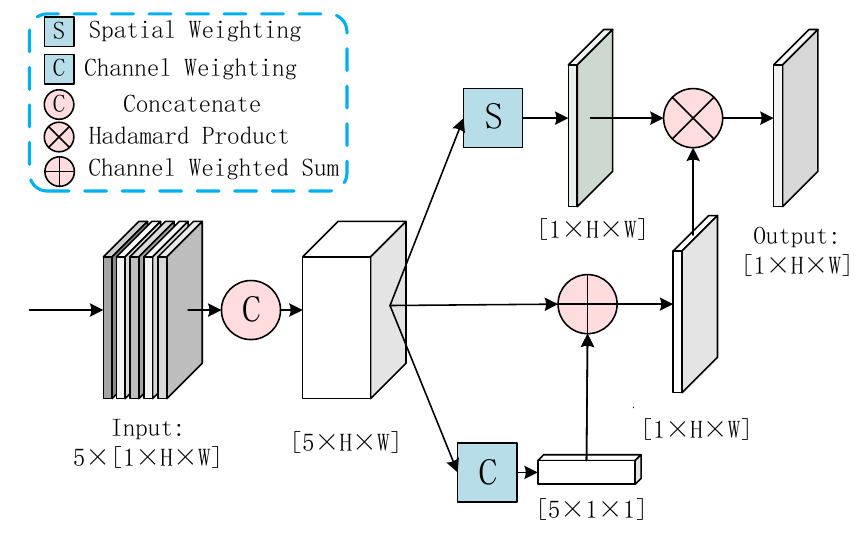}
	\caption{Pseudo Pixel-level Weighting module. The input of the PPW module is the multiscale feature and the output is the final edge. "5" represents the number of multi-scale feature maps, $H$ and $W$ represent the height and width of the image, which are also the height and width of the input and output of the PPW module.}
    \label{fig-PPW}
\end{figure}

In the second feature fusion stage, we propose a Pseudo Pixel-level Weighting (PPW) module that decomposes the weight of each pixel into spatial and channel components as depicted in Fig.~\ref{fig-PPW}. To minimize computation cost, we use a $1\times1$ convolution to directly calculate the product of channel weight and multi-scale edges. The spatial weight is calculated using a spatial weighting module consisting of three $3\times3$ convolution layers and a softmax activation function, similar to the CoFusion~\cite{huan2021unmixing}. Since the spatial weighting module only accounts for spatial weight, fewer channels are required. Experimental results show that spatial weighting of PPW is able to achieve comparable accuracy with CoFusion using only 25\% of the number of channels.

The final prediction of each pixel $P_{ij}$ can be calculated using the PPW module input $X_{ijk}$ as follows:
\begin{equation}
    \begin{split}
        P_{ij} &= PPW(X_{ijk}) \\
               &= \sum_{i=1}^{L}(X_{ijk}\times W_{ijk}) \\
               &= \sum_{i=1}^{L}(X_{ijk}\times Wc_i \times Ws_{jk})
    \end{split}
\end{equation}
Here, \textit{L} represents the number of multi-scale edges, and \textit{Wc} and \textit{Ws} represent channel weight and spatial weight, respectively. \textit{X} and \textit{P} denote the input and output of the PPW module.

The PPW module bears some resemblance to CBAM~\cite{DBLP:conf/eccv/WooPLK18}, but there are two primary differences. 
Firstly, the two modules serve different purposes. The primary objective of PPW is to assign weights to the fusion of multi-scale features to obtain higher quality single-channel images. In contrast, CBAM's primary objective is to selectively emphasize informative features in their channel attention mechanism.
Secondly, there are differences in implementation. Specifically, in PPW, a $1\times 1$ convolution is used directly to perform channel weighting, and channels are reduced as early as possible in order to decrease module complexity.

\subsection{Dynamic focal loss} \label{section_DFL}

Edge and non-edge pixels are extremely imbalanced in images, thus Weighted Cross-Entropy loss (WCE) is widely employed in edge detection, which is formulated as
\begin{equation}
	WCE\left(p_{i}, y_{i}\right)=\left\{
	\begin{array}{cc}
		-\alpha \log \left(p_{i}\right)  &  { if }\ \mathrm{y}_{i}=1 \\
		-\beta \log \left(1-p_{i}\right) &  { if }\ \mathrm{y}_{i}=0 \\
		0								 &  { otherwise }
	\end{array}\right.
\end{equation}
where \textit{p} denotes the final edge prediction and \textit{y} represents the ground truth. $\alpha=\lvert Y_{+} \rvert/\lvert Y \rvert$ and $\beta =\lambda\cdot\lvert Y_{-} \rvert/\lvert Y \rvert$. $\lvert Y_{+} \rvert$ and $\lvert Y_{-} \rvert$ are used to represents the number of edge and non-edge, separately. $\lambda$ controls the weight of positive over negative samples. $\lvert Y \rvert=\lvert Y_{+} \rvert + \lvert Y_{-} \rvert$. Due to the inconsistency of annotations among different annotators, a threshold $\gamma$ is introduced for loss computation.  For pixel \emph{i}, it will be regarded as edge if the ground truth $y_{i}$ is more than $\gamma$, and we define the ground truth $y_{i}=1$. Pixel \textit{i} will be regarded as no-edge if $y_{i}=0$. $\lambda$ controls the weight of edge over non-edge. 

WCE solves the unbalance between positive and negative samples in edge detection by a balance coefficient successfully while ignoring the problem of hard samples. Hard samples are those pixels easily misclassified, which determines the quality of the edge map. 
Hard samples are common in dense prediction tasks. In view of this problem, Lin \etal proposed Focal Loss (FL)~\cite{lin2017focal}
\begin{equation}
	F L(p_{i}, y_{i})=\left\{\begin{array}{cc}
		-\alpha\omega \log (p_{i})    &  { if } \ y_{i}=1 \\
		-\beta \omega \log (1-p_{i}) &  { otherwise } .
	\end{array}\right.
 \label{eq-fl}
\end{equation}
where
\begin{equation}
    \omega=\left\{\begin{array}{cc}
        (1-p_{i})^\gamma  &  { if } \ y_{i}=1 \\
        p_{i}^\gamma    &  { otherwise. }
    \end{array}\right.
\end{equation}

Compared with WCE, FL contains a new weight factor $\omega$, which utilizes the difference between the predicted result and the ground truth to weigh the sample and adjusts the weight flexibly by a hyper-parameter $\gamma$. Therefore, the model is guided to pay more attention to hard samples. The effectiveness of FL has been widely demonstrated~\cite{tian2019fcos,carion2020end}. 

However, FL is suffering from confusion in the early stage of model training. Because some modules of the model are randomly initialized, the difference between the prediction and the ground truth at the beginning of training cannot truly reflect the distribution of hard samples. At this point, using FL to focus on fake hard samples is detrimental to the learning of real hard samples. So, we propose Dynamic Focal Loss (DFL) to optimize this situation. DFL is formulated as
\begin{equation}
	DFL(p_{i}, y_{i})=\left\{\begin{array}{ccc}
		-\alpha \omega' \log (p_{i})         &  { if }\ y_{i}=1 \\
		-\beta \omega' \log (1-p_{i})   &  { if }\ y_{i}=0 \\
		0                              &  { otherwise. }
	\end{array}\right.
\end{equation}
where

\begin{equation}
    \omega'=\left\{\begin{array}{cc}
        \frac{\mu+\epsilon(1-p_{i})^\gamma}{\epsilon+\mu}  &  { if } \ y_{i}=1 \\
        \frac{\mu+\epsilon(p_{i})^\gamma}{\epsilon+\mu}    &  { otherwise. }
    \end{array}\right.
    \label{eq-omega'}
\end{equation}
$\epsilon$ represents the current epoch and starts from 0. The effect of hyper-parameters $\gamma$ and $\mu$ will be discussed in ablation study.

Our contribution is based on the fact that the output of the model can gradually reflect the correct distribution of the hard samples as the training progresses. Hence, our confidence margin in the prediction should rise gradually.
Here we represent this process as an exponential function $\epsilon/(\epsilon+\mu)$. 

The experimental results show that this simple dynamic confidence setting can better define hard samples and guide the model to converge in a more correct direction.

\section{Experiments} \label{sec:experiments}

In this section, we first introduce the datasets and implementation details of the experiment, then compare our method with the State-of-the-art methods, and finally verify the effectiveness of each module in our method through ablation experiments and Visual Analysis.

\subsection{Experimental datasets}

The performance of our proposed CTFN is validated on three benchmark datasets (BSDS500~\cite{arbelaez2010contour}, NYUDv2~\cite{silberman2012indoor}, and BIPEDv2~\cite{poma2020dense}) and compare with the previous state-of-the-art methods. \textbf{BSDS500} is the most popular dataset for edge detection, including 200 training images, 100 validation images, and 200 test images, each of which has 4-9 corresponding annotation results. In the experiment, a total of 300 images in the training and validation sets are used to train the model, and then the model is evaluated on the test set. \textbf{NYUDv2} is an indoor scene semantic segmentation dataset, the edge ground truth is generated from segmentation maps, it containing 1449 groups of carefully annotated RGB and depth images, and each group image has one annotation result. We use 795 images to train the model and evaluate the model on the rest images. \textbf{BIPEDv2} is recently proposed by Soria \etal~\cite{soria2021bipedv2}, which is the second version of~\cite{poma2020dense}. This dataset contains 250 carefully annotated high-resolution Barcelona Street View images. There are 200 images for training and validation and 50 images for testing. For a fair comparison, we use the same data augmentation method as RCF~\cite{liu2019richer} for BSDS500 and NYUDv2, and the same as DexiNed~\cite{poma2020dense} for BIPEDv2.

\subsection{Implementation Details.}\label{sec:Implementation-details}
Our method is implemented on the PyTorch library. All the experiments are conducted on an NVIDIA GeForce 2080Ti GPU with 11GB memory. The backbone of CTFN is initialized on ImageNet and the rest module is randomly initialized. The threshold $\gamma$ is set to 0.3 for BSDS500. And $\gamma$ is useless for NYUDv2 and BIPEDv2, due to the ground truth being binary annotations. $\lambda$ is set to 1.1 for BSDS500 and BIPEDv2, 1.2 for NYUDv2. SGD optimizer is adopted. A standard Non-Maximum Suppression (NMS) is performed to produce the final edge maps before the quantitative evaluation. F-measure is utilized as the quality evaluation standard of the generated edge map:
\begin{equation}
	F_{-} measure=\frac{2^{*} P^{*} R}{P+R}
\end{equation}
where \emph{P} represents the accuracy rate and \emph{R} represents the recall rate. 

Due to local correlation, the edges obtained by deep learning-based methods are actually edge probability maps, even after NMS processing. So we need to choose a threshold to binarize the edges.
There are two options for the threshold used to binarize the edges. One is to find an optimal threshold value for each image (OIS), and another is to use the same optimal threshold for the whole dataset (ODS). 
The maximum tolerance allowed for correct matches between edge predictions and ground truth annotations is set to 0.0075 for BSDS500 and BIPEDv2, 0.011 for NYUDv2. More experimental details can be referred to previous work~\cite{huan2021unmixing,poma2020dense}.

\subsection{Comparison with the State-of-the-art Methods}

\subsubsection{Performance on BSDS500}

We compare CTFN with recent deep learning based edge detection methods on BSDS500, and the results are summarized in Table~\ref{tab-bsds-vgg}. To be fair, all methods are based on VGG16. In terms of accuracy, CTFN, BAN, and FCL are significantly better than other methods. CTFN and BAN have advantages in ODS and OIS respectively and are all slightly better than FCL. In terms of the number of parameters, CTFN, RCF, and CAT are fewer than other methods. \emph{P'} can reflect this advantage more intuitively. Analyzing parameters except for the VGG16 backbone, CTFN is 1/6 of BDCN, 1/9 of BAN, and 1/18 of FCL. In terms of the amount of calculation, FLOPs of CTFN is only 0.6G more than RCF, ranking second. FLOPs of CTFN is 25\% less than BDCN, and nearly half less than FCL or BAN. In short, CTFN's accuracy is comparable to that of the state-of-the-art methods, while the number of parameters and computation cost is far superior to them. Additionally, BAN utilizes several tricks like combined loss function, edge map merging~\cite{xie2017holistically}, and two-stage training, while our CTFN is trained in an end-to-end manner.

\begin{table}[htbp]
	\begin{center}
    \caption{Comparison between CTFN with other VGG16-based methods on BSDS500 dataset. \emph{P} represents the total number of model parameters, and \emph{P'} represents the number of parameters except for the backbone. The best two results are highlighted in \cred{red} and \cblue{blue}, respectively, and same for other tables.}
    \label{tab-bsds-vgg}
     \begin{tabular}{ccccc}
    		\toprule
    		Method  & ODS$\uparrow$    & OIS$\uparrow$  & P/P'(M)$\downarrow$ &   FLOPs(G)$\downarrow$ \\
    		\midrule
    		RCF		&0.798          &0.817         &\textcolor{red}{14.8}/\textcolor{red}{0.1}	&  	\textcolor{red}{75.3}\\
    		CATS	&0.806          &0.821         &\textcolor{red}{14.8}/\textcolor{red}{0.1}	&	78.3\\
    		BDCN	&0.806          &0.822         & 16.3/1.6			&	103.4\\
    		FCL 	&\textcolor{blue}{0.816}          &0.831         & 16.5/1.8			&	134.4\\
    		BAN	 	&\textcolor{blue}{0.816}     &\textcolor{red}{0.834}   & \textcolor{blue}{15.6}/\textcolor{blue}{0.9}			&	142.2\\
    		CTFN   &\textcolor{red}{0.817}    &\textcolor{blue}{0.832}      &\textcolor{red}{14.8}/\textcolor{red}{0.1}	&	\textcolor{blue}{75.9}\\
    		\bottomrule
    	\end{tabular}
    \end{center}
\end{table}

\begin{figure}[htbp]
	\centering
	\includegraphics[width=0.6\linewidth]{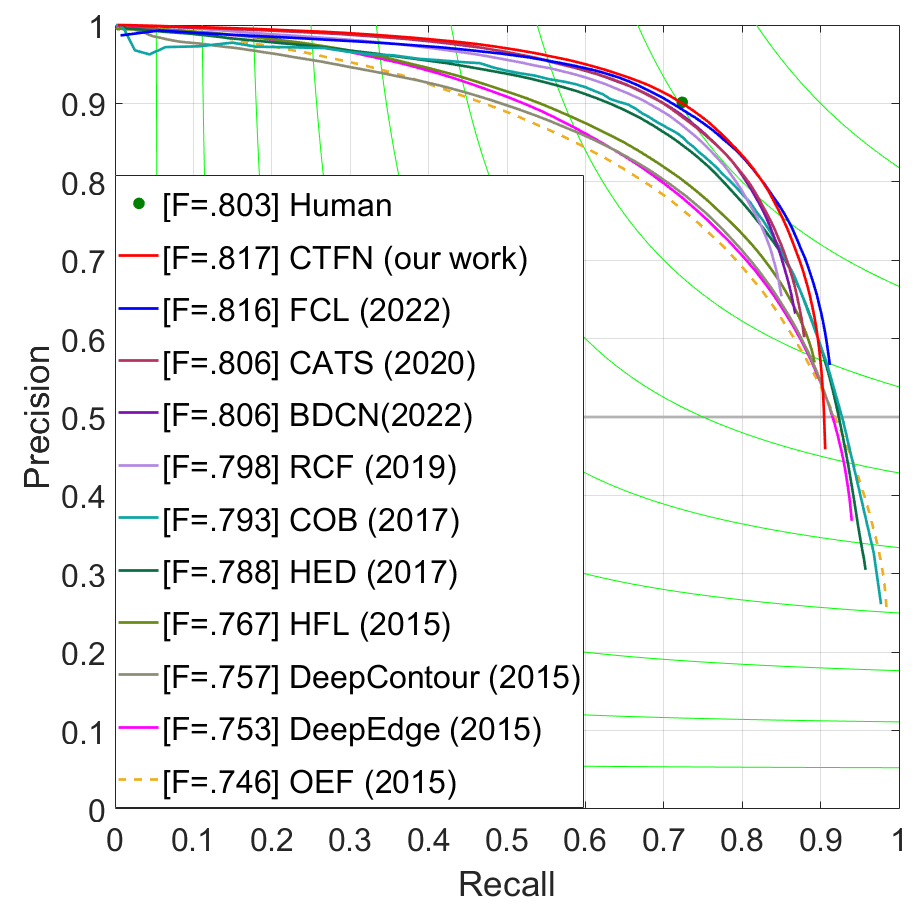}
	\caption{Precision–Recall curves of CTFN compared with other existing works on BSDS500 dataset.} 
	\label{fig-pr-bsds}
\end{figure}

\begin{figure*}[htbp]
	\centering
  \hspace{-0.2cm}
	\subfigure[image]{
		\begin{minipage}[t]{0.15\textwidth}
			\centering
   			\includegraphics[width=\textwidth]{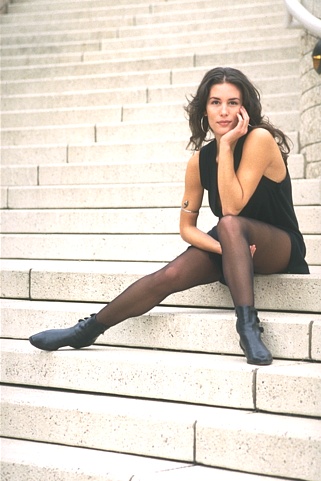}\\
			\vspace{0.05cm}
   			\includegraphics[width=\textwidth]{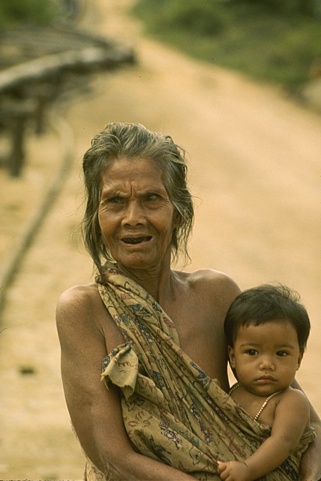}\\
			\vspace{0.05cm}
   			\includegraphics[width=\textwidth]{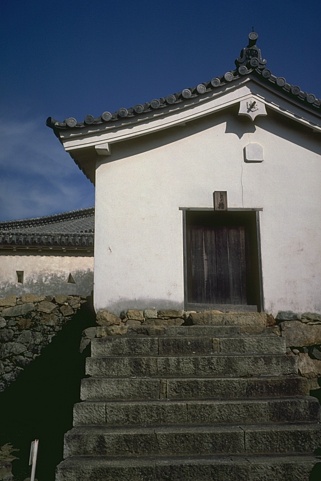}\\
		\end{minipage}%
	}
 \hspace{-0.2cm}
	\subfigure[GT]{
		\begin{minipage}[t]{0.15\textwidth}
			\centering
   			\includegraphics[width=\textwidth]{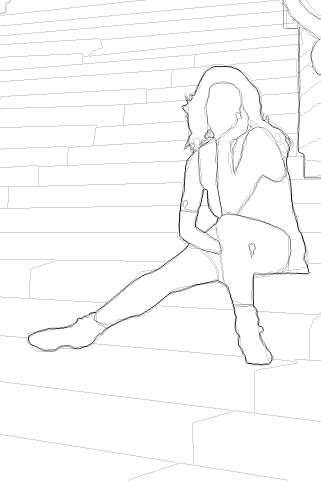}\\
			\vspace{0.05cm}
   			\includegraphics[width=\textwidth]{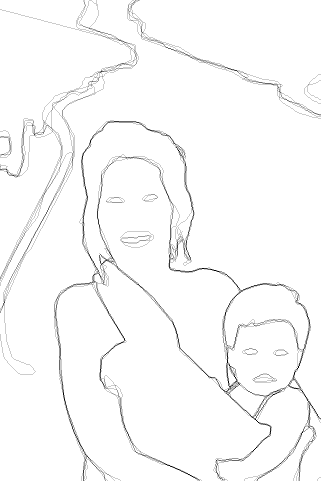}\\
			\vspace{0.05cm}
   			\includegraphics[width=\textwidth]{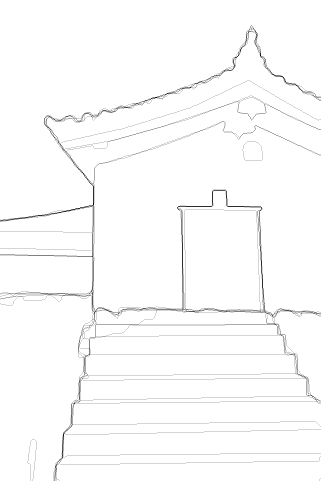}\\
		\end{minipage}%
	}
  \hspace{-0.2cm}
	\subfigure[HED]{
		\begin{minipage}[t]{0.15\textwidth}
			\centering
   			\includegraphics[width=\textwidth]{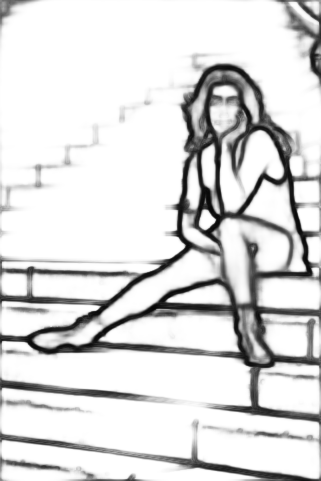}\\
			\vspace{0.05cm}
   			\includegraphics[width=\textwidth]{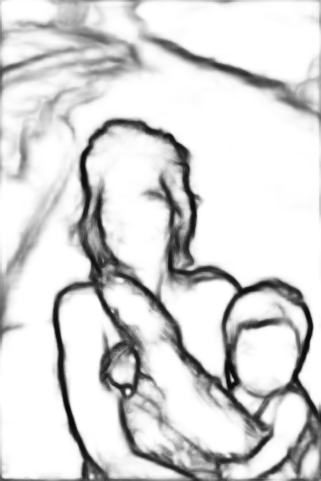}\\
			\vspace{0.05cm}
   			\includegraphics[width=\textwidth]{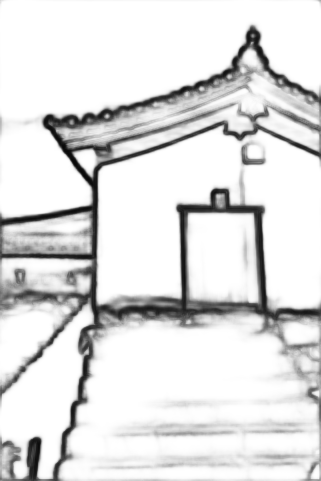}\\
		\end{minipage}%
	}
  \hspace{-0.2cm}
	\subfigure[RCF]{
		\begin{minipage}[t]{0.15\textwidth}
			\centering
   			\includegraphics[width=\textwidth]{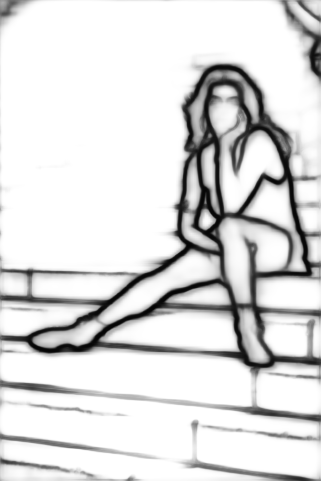}\\
			\vspace{0.05cm}
   			\includegraphics[width=\textwidth]{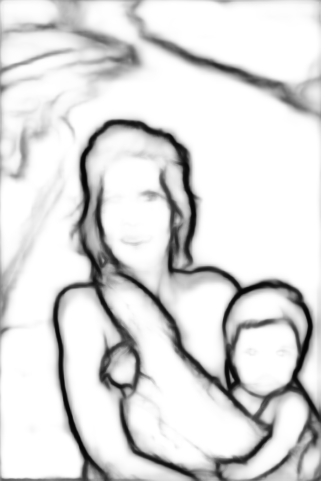}\\
			\vspace{0.05cm}
   			\includegraphics[width=\textwidth]{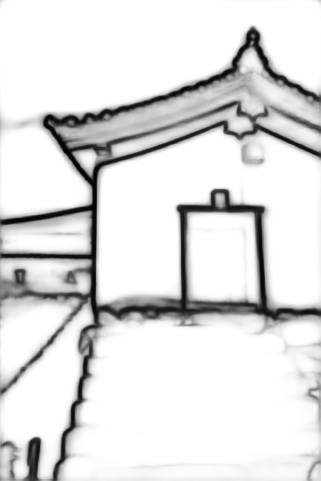}\\
		\end{minipage}%
  }
    \hspace{-0.2cm}
	\subfigure[BDCN]{
		\begin{minipage}[t]{0.15\textwidth}
			\centering
   			\includegraphics[width=\textwidth]{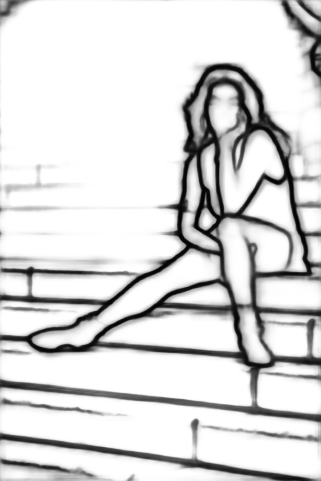}\\
			\vspace{0.05cm}
   			\includegraphics[width=\textwidth]{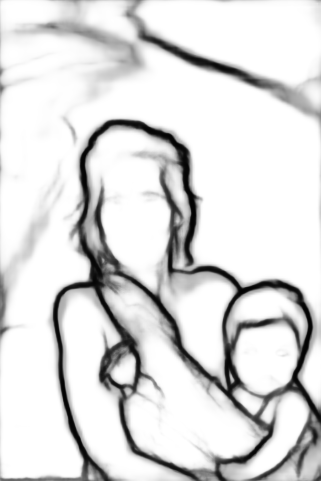}\\
			\vspace{0.05cm}
   			\includegraphics[width=\textwidth]{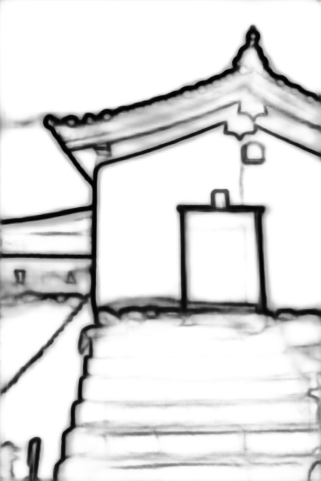}\\
		\end{minipage}%
  }
    \hspace{-0.2cm}
	\subfigure[CTFN]{
		\begin{minipage}[t]{0.15\textwidth}
			\centering
   			\includegraphics[width=\textwidth]{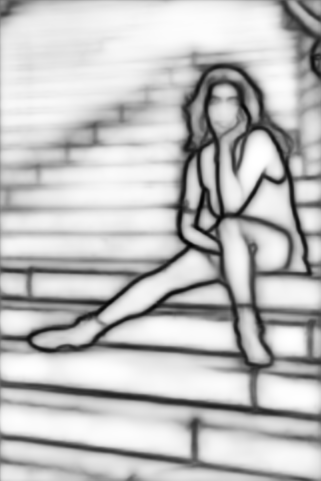}\\
			\vspace{0.05cm}
   			\includegraphics[width=\textwidth]{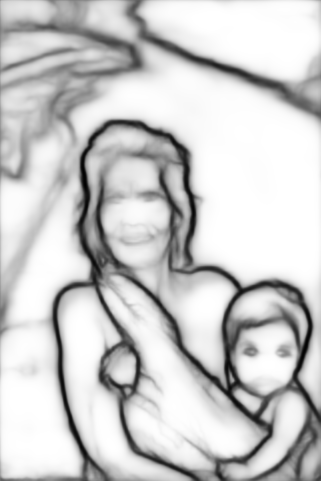}\\
			\vspace{0.05cm}
   			\includegraphics[width=\textwidth]{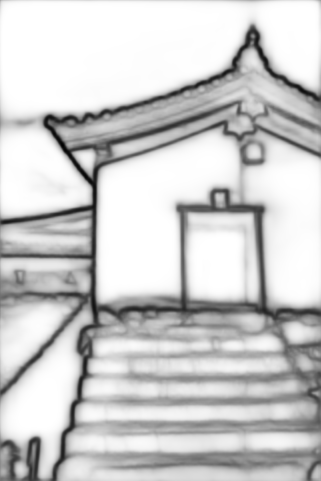}\\
		\end{minipage}%
	}
 
	\caption{Comparison of edge detection results on BSDS dataset. All the results are raw edge maps before Non-Maximum Suppression.}\label{fig-bsds}
	
\end{figure*}

A larger scale accuracy comparison is shown in Table~\ref{tab_bsds}. CTFN is compared with prior edge detection methods, including both traditional methods and deep learning methods. As shown in Table~\ref{tab_bsds}, CTFN reports ODS 0.817 on BSDS500, which is around 2\% higher than the baseline method RCF. CTFN outperforms most existing HED-based methods and UNet-based methods with a large gap. Precision-Recall curves are presented in Fig.~\ref{fig-pr-bsds}. As can be seen from the PR curves, when recall rate is less than 0.8, CTFN maintains a significant advantage in accuracy. In addition, we visualized the prediction results of images in BSDS500, and the comparison results are shown in Fig.~\ref{fig-bsds}. Visually, CTFN is significantly superior to other methods.

\begin{table}[htbp]
	\begin{center}
    \caption{\centering{Comparison with other methods on BSDS500 dataset.}}
    \label{tab_bsds}
	\begin{tabular}{lclll}
		\toprule
		Method 			&  Pub.'year' 	& ODS & OIS\\
		\midrule
		Canny~\cite{canny1986computational}		& 	PAMI’86		& 0.611	& 	0.676\\
		Pb~\cite{martin2004learning} 		    & 	PAMI’04		& 0.672 & 	0.695\\
		Felz-Hutt~\cite{ferrari2007groups}		& 	IJCV’04  	& 0.61  & 	0.640\\
		gPb-owt-ucm~\cite{arbelaez2010contour}	& 	PAMI’10   	& 0.726 & 	0.757\\
		SCG~\cite{xiaofeng2012discriminatively}	& 	NeurIPS’12	& 0.739	& 	0.758\\
		Sketch Tokens~\cite{lim2014learned}      & 	CVPR’13   	& 0.727 & 	0.746\\
		SE~\cite{dollar2014fast}   			    & 	PAMI’14	  	& 0.746 & 	0.767\\
		OEF~\cite{hallman2015oriented}	   		& 	PAMI’14   	& 0.746 & 	0.770\\
		MES~\cite{sironi2015projection}   		& 	ICCV’15   	& 0.756 & 	0.776\\
		\midrule
		DeepEdge~\cite{bertasius2015deepedge}   	& 	CVPR’15	   	& 0.753 & 	0.772\\
		DeepContour~\cite{shen2015deepcontour}   & 	CRVR’15	   	& 0.757 & 	0.776\\
		Deep Boundary~\cite{kokkinos2015pushing} & 	ICLR’15   	& 0.813 & 	0.831\\
		CEDN~\cite{yang2016object}   		    & 	CVPR’16    	& 0.788 & 	0.804\\
		COB~\cite{martin2004learning}   			& 	ECCV’16   	& 0.793 & 	0.820\\
		DCD~\cite{liao2017deep}   			    & 	ICME’17   	& 0.799 & 	0.817\\
		CED~\cite{wang2018deep}	   		        & 	CVPR’17   	& 0.803 & 	0.820\\
		HED~\cite{xie2017holistically}	   	    & 	IJCV’17	   	& 0.788 & 	0.807\\
		LPCB~\cite{deng2018learning}		        & 	ECCV’18	   	& 0.800 & 	0.816\\
		RCF~\cite{liu2019richer}	   	            & 	PAMI’19   	& 0.798 & 	0.817\\
		DexiNed~\cite{poma2020dense}	   	        & 	WACV’20	   	& 0.729	& 	0.745\\
		DSCD~\cite{deng2020deep}   		        & 	ACMMM’20   	& 0.802 & 	0.817\\
        BAN~\cite{gao2021bottom}	   		        & 	IJCAI’20   	& \textcolor{blue}{0.816}	& 	\textcolor{red}{0.834}\\
		PiDiNet~\cite{su2021pixel}	   	        & 	ICCV’21   	& 0.807	& 	0.823\\
		CATS~\cite{huan2021unmixing}   	        & 	PAMI’21	   	& 0.806 & 	0.821\\
		BDCN~\cite{he2022bi-directional}   	    & 	PAMI’22	   	& 0.806 & 	0.822\\
		FCL~\cite{xuan2022fcl}   		        & 	NN’22   	& \textcolor{blue}{0.816}	& 	0.831\\
		\midrule
		CTFN	   		                        & 	  -{}-   	& \textcolor{red}{0.817} & 	\textcolor{blue}{0.832}\\
		\bottomrule
	\end{tabular}
    \end{center}
\end{table}

\subsubsection{Performance on NYUDv2 and BIPEDv2}

\begin{table}[htbp]
		\begin{center}
        \caption{Quantitative comparison on NYUDv2 dataset with CTFN and other methods considered for the evaluation.}
		\label{tab-nyud}
		\begin{tabular}{lcll}
			\hline
			\multicolumn{2}{c}{Method}  & ODS & OIS\\
			\toprule
			OEF~\cite{hallman2015oriented} &	RGB	& 0.651	 & 0.667 \\
			SE~\cite{dollar2014fast}	& 	RGB 	& 0.695		& 0.708\\
			\midrule
			
			\multirow{3}*{HED~\cite{xie2017holistically}}	& 	RGB		& 	0.720	& 	0.734\\
			&   HHA		& 0.682		& 	0.695\\
			&RGB+HHA	& 	0.746	& 0.731\\
			\midrule
			\multirow{3}*{RCF~\cite{liu2019richer}}	& 	RGB		& 0.729		& 0.742\\
			&	HHA		& 0.705		& 	0.715\\
			&	RGB+HHA	& 0.757		& 	0.771\\
			\midrule
			
			\multirow{3}*{BDCN~\cite{he2022bi-directional}}	& 	RGB		& 	0.748	& 	0.763\\
			&	HHA		& 	0.707	& 	0.719\\
			&RGB+HHA	& \textcolor{blue}{0.765}		& \textcolor{blue}{0.781}\\
			\midrule
			\multirow{3}*{CTFN}		& 	RGB		& 0.752		& 0.767\\
			&	HHA		& 	0.712	& 	0.721\\
			& RGB+HHA	& 	\textcolor{red}{0.769}	& 	\textcolor{red}{0.785}\\
			\bottomrule
		\end{tabular}
        \end{center}
	\end{table}

\begin{figure*}[htbp]
	\centering
	\subfigure[image]{
		\begin{minipage}[t]{0.19\textwidth}
			\centering
			\includegraphics[width=\textwidth]{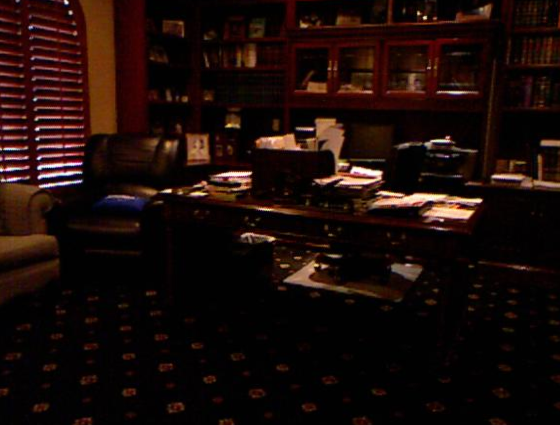}\\
			\vspace{0.05cm}
			\includegraphics[width=\textwidth]{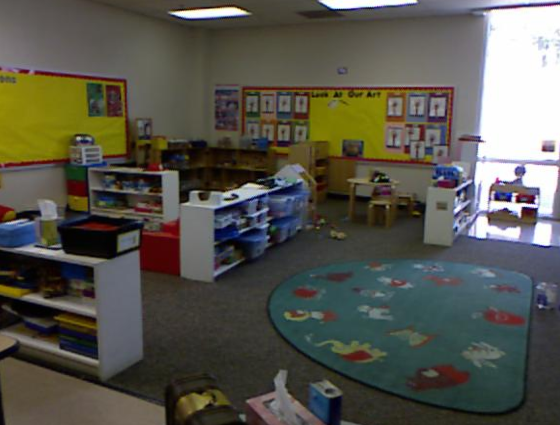}\\
			\vspace{0.05cm}
		\end{minipage}%
	}
	\hspace{-0.2cm}
	\subfigure[GT]{
		\begin{minipage}[t]{0.19\textwidth}
			\centering
			\includegraphics[width=\textwidth]{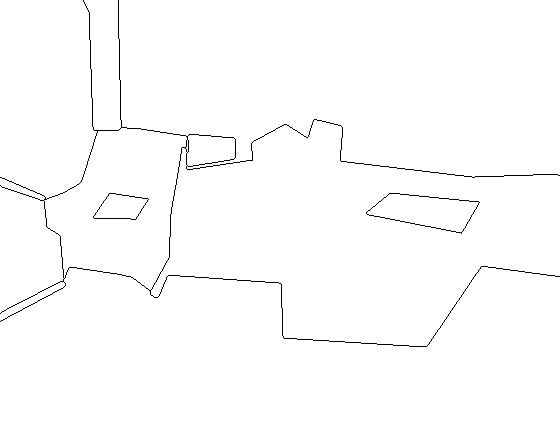}\\
			\vspace{0.05cm}
			\includegraphics[width=\textwidth]{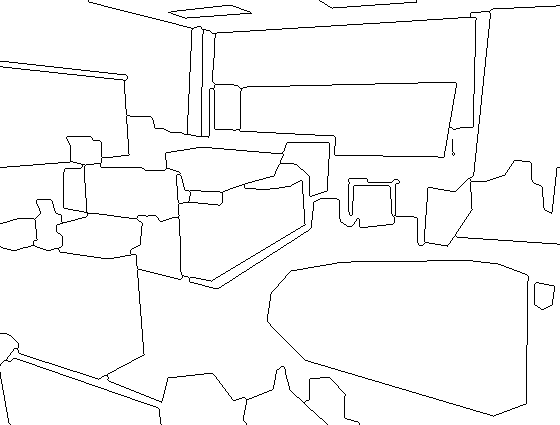}\\
			\vspace{0.05cm}
		\end{minipage}%
	}
	\hspace{-0.2cm}
	\subfigure[RCF~\cite{liu2019richer}]{
		\begin{minipage}[t]{0.19\textwidth}
			\centering
			\includegraphics[width=\textwidth]{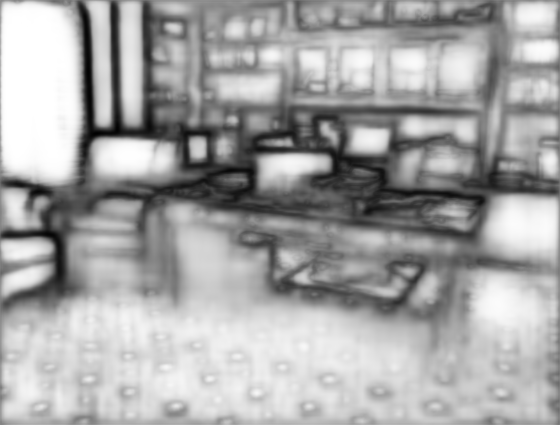}\\
			\vspace{0.05cm}
			\includegraphics[width=\textwidth]{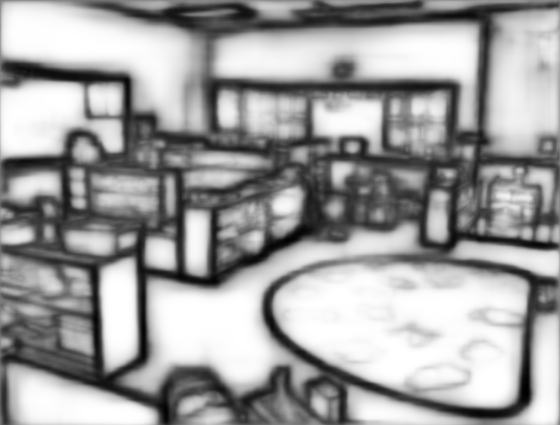}\\
			\vspace{0.05cm}
		\end{minipage}%
	}
	\hspace{-0.2cm}
	\subfigure[BDCN~\cite{he2022bi-directional}]{
		\begin{minipage}[t]{0.19\textwidth}
			\centering
			\includegraphics[width=\textwidth]{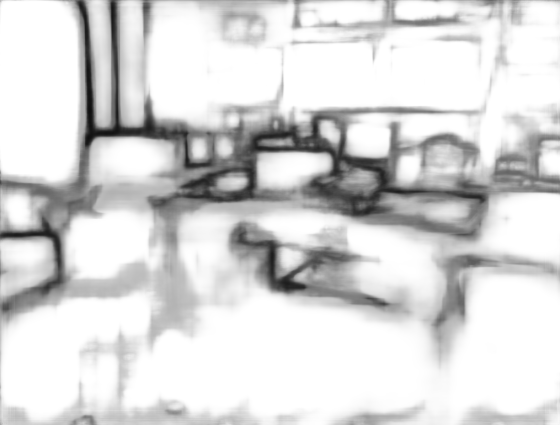}\\
			\vspace{0.05cm}
			\includegraphics[width=\textwidth]{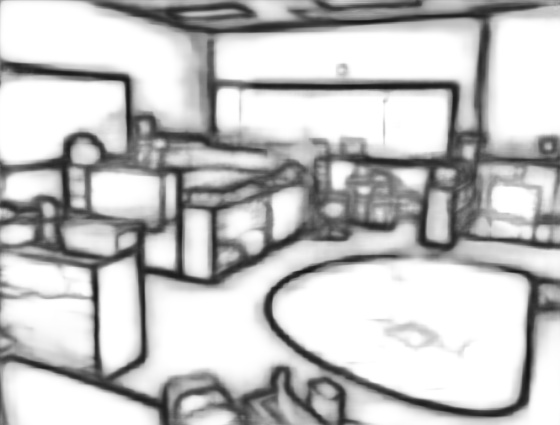}\\
			\vspace{0.05cm}
		\end{minipage}%
	}
	\hspace{-0.2cm}
	\subfigure[CTFN]{
		\begin{minipage}[t]{0.19\textwidth}
			\centering
			\includegraphics[width=\textwidth]{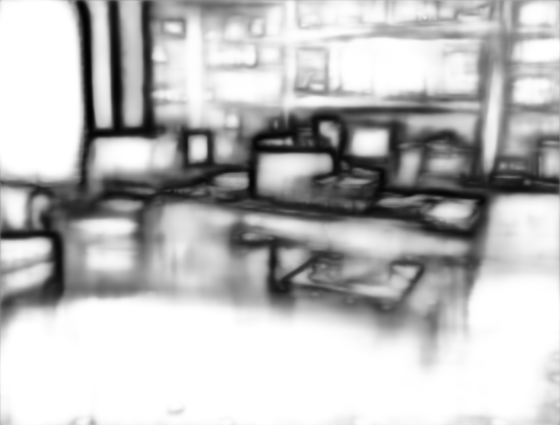}\\
			\vspace{0.05cm}
			\includegraphics[width=\textwidth]{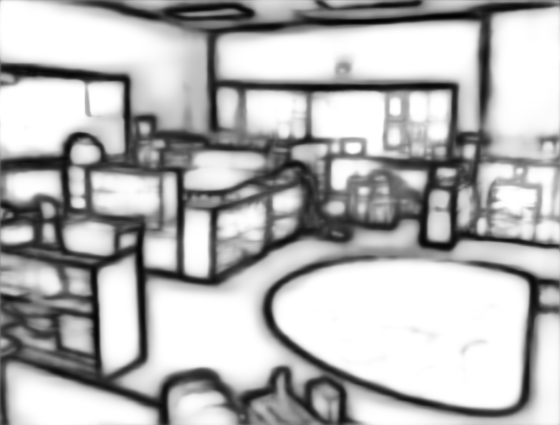}\\
			\vspace{0.05cm}
		\end{minipage}%
	}

	\caption{Comparison of edge detection results on NYUD dataset. All the results are raw edge maps before Non-Maximum Suppression.}\label{fig-nyud}
	
\end{figure*}

NYUDv2 has three types of inputs, i.e., RGB, HHA, and RGB-HHA. Following previous works~\cite{liu2019richer,he2022bi-directional}, we perform experiments on the three types data. The results of RGB-HHA are obtained by averaging the edges detected on RGB and HHA. Table~\ref{tab-nyud} shows the comparison of our method with several recent approaches. Our method outperforms the baseline RCF by 0.4\% on RGB-HHA. A comparison of edge detection results on NYUDv2 is shown in Fig.~\ref{fig-nyud}. From left to right are input images, ground truth, and the results of RCF, BDCN, and CTFN. The edges generated by RCF contain many textures that should have belonged to the background. While BDCN, on the other hand, loses many edges. The results of CTFN were significantly better than those from RCF and BDCN. 

\begin{table}[htbp]
    \begin{center}
     \caption{Quantitative comparison on BIPEDv2 dataset with CTFN and other methods considered for the evaluation.}
    \label{tab-biped}
    \begin{tabular}{lll}
        \toprule
        Method 		& ODS 	& OIS\\
        \midrule
        HED~\cite{xie2017holistically}			& 0.883 &	0.890\\
        RCF~\cite{liu2019richer} 		& 0.886 &	0.892\\
        BDCN~\cite{he2022bi-directional} 		& \textcolor{blue}{0.894} &	\textcolor{blue}{0.899}\\
        DexiNed-a~\cite{poma2020dense} 	& 0.893 &	0.897\\
        DexiNed-f~\cite{poma2020dense} 	& \textcolor{red}{0.895} &	\textcolor{red}{0.900}\\
        FCL~\cite{xuan2022fcl}	 		& \textcolor{red}{0.895} &	\textcolor{red}{0.900}\\
        CTFN 		& \textcolor{blue}{0.894} &	\textcolor{blue}{0.899}\\
        \bottomrule
    \end{tabular}
    \end{center}
\end{table}

\begin{figure*}[htbp]
	\centering
	\subfigure[image]{
		\begin{minipage}[t]{0.22\textwidth}
			\centering
			\includegraphics[width=\textwidth]{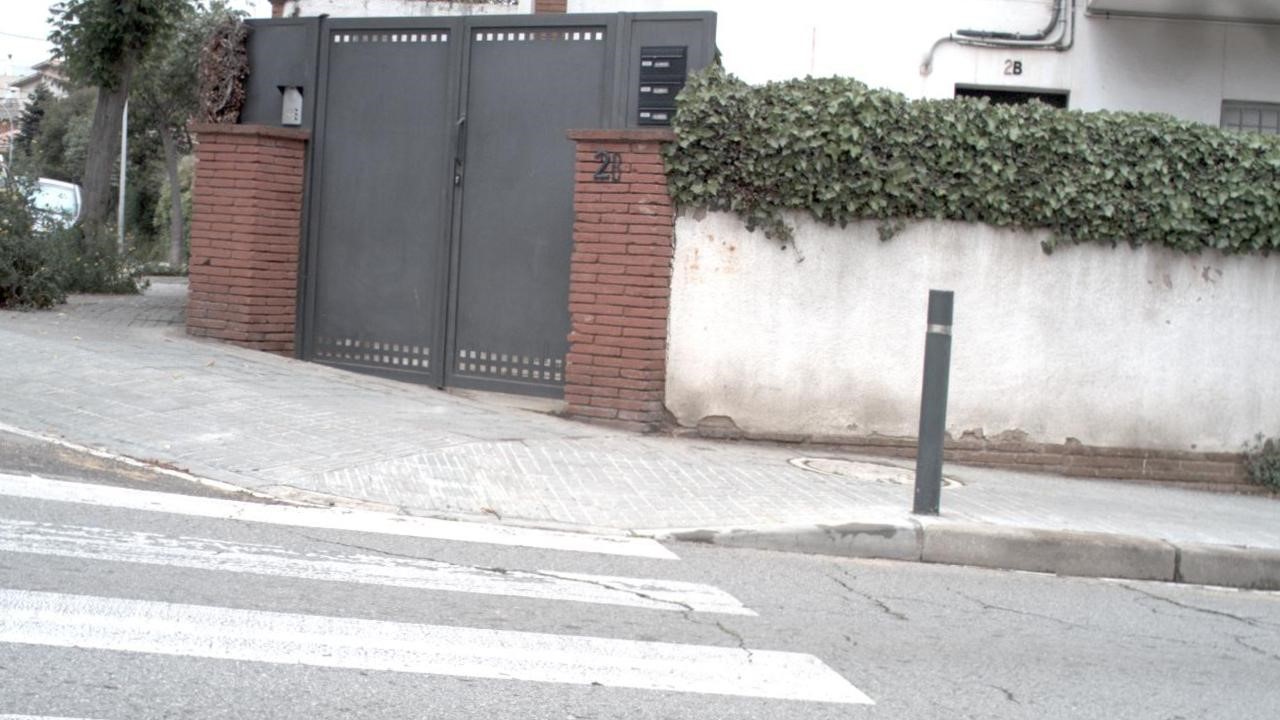}\\
			\vspace{0.05cm}
			\includegraphics[width=\textwidth]{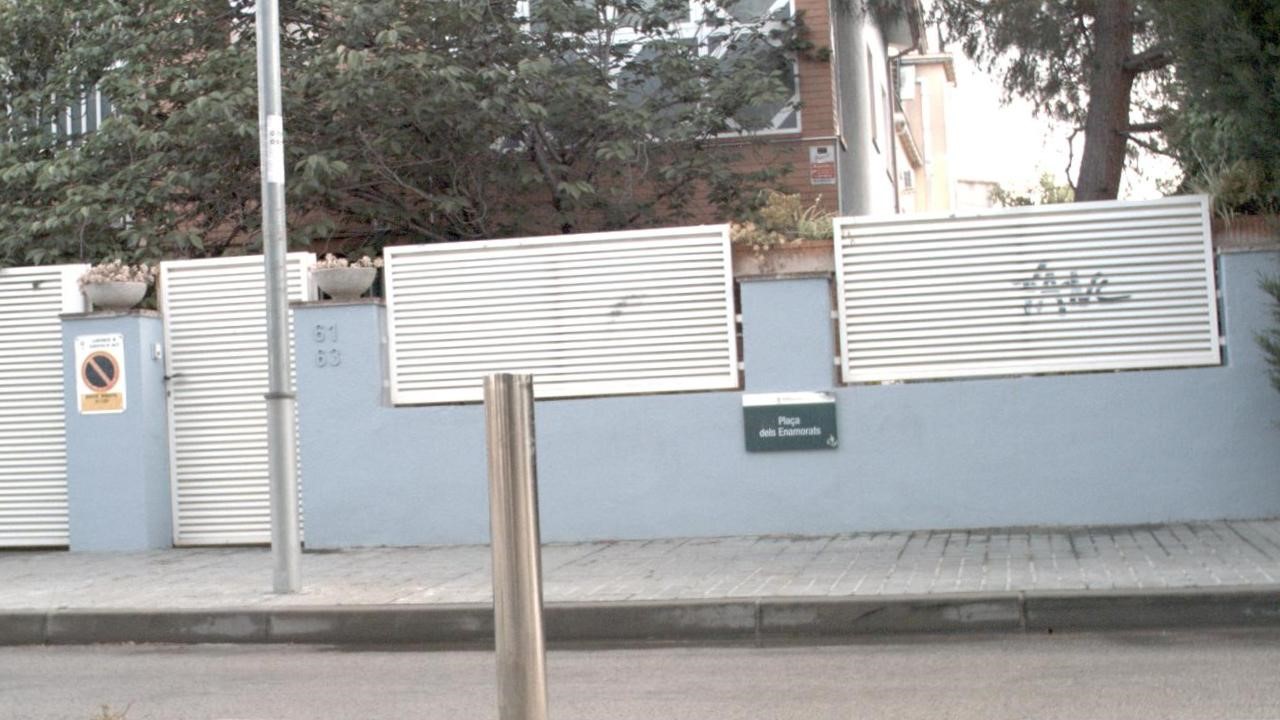}\\
			\vspace{0.05cm}
		\end{minipage}%
	}
	\subfigure[GT]{
		\begin{minipage}[t]{0.22\textwidth}
			\centering
			\includegraphics[width=\textwidth]{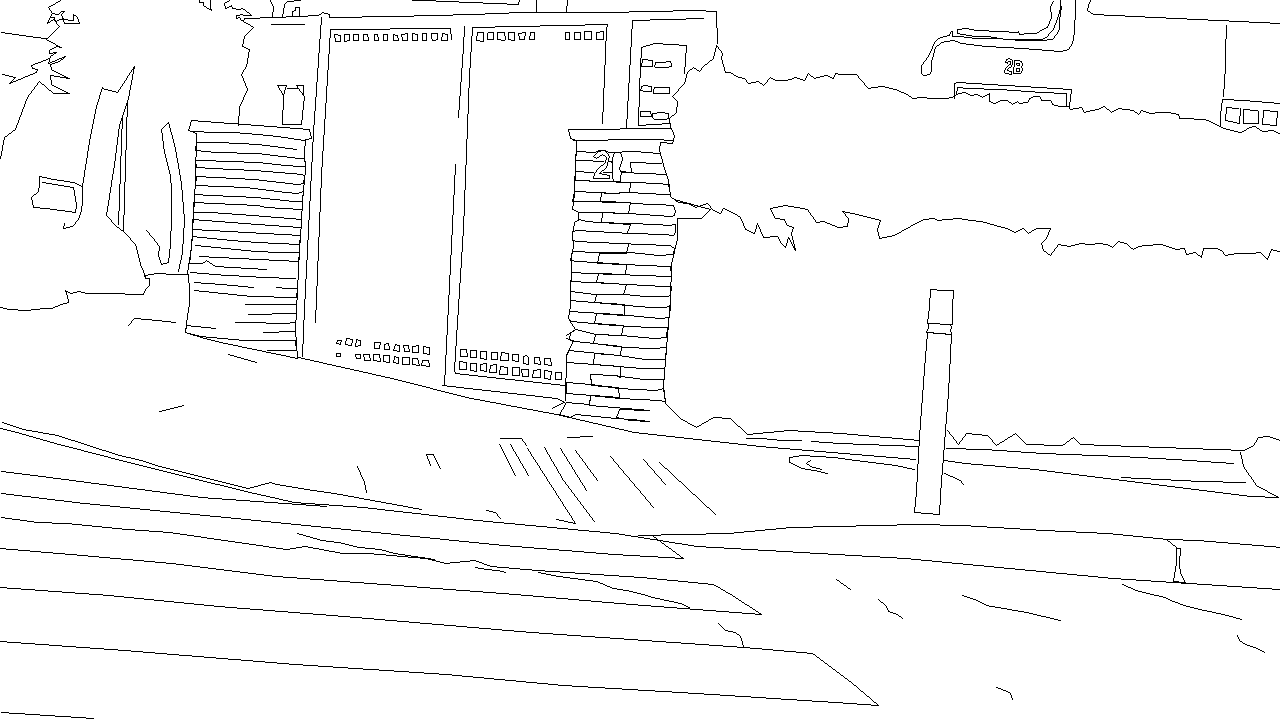}\\
			\vspace{0.05cm}
			\includegraphics[width=\textwidth]{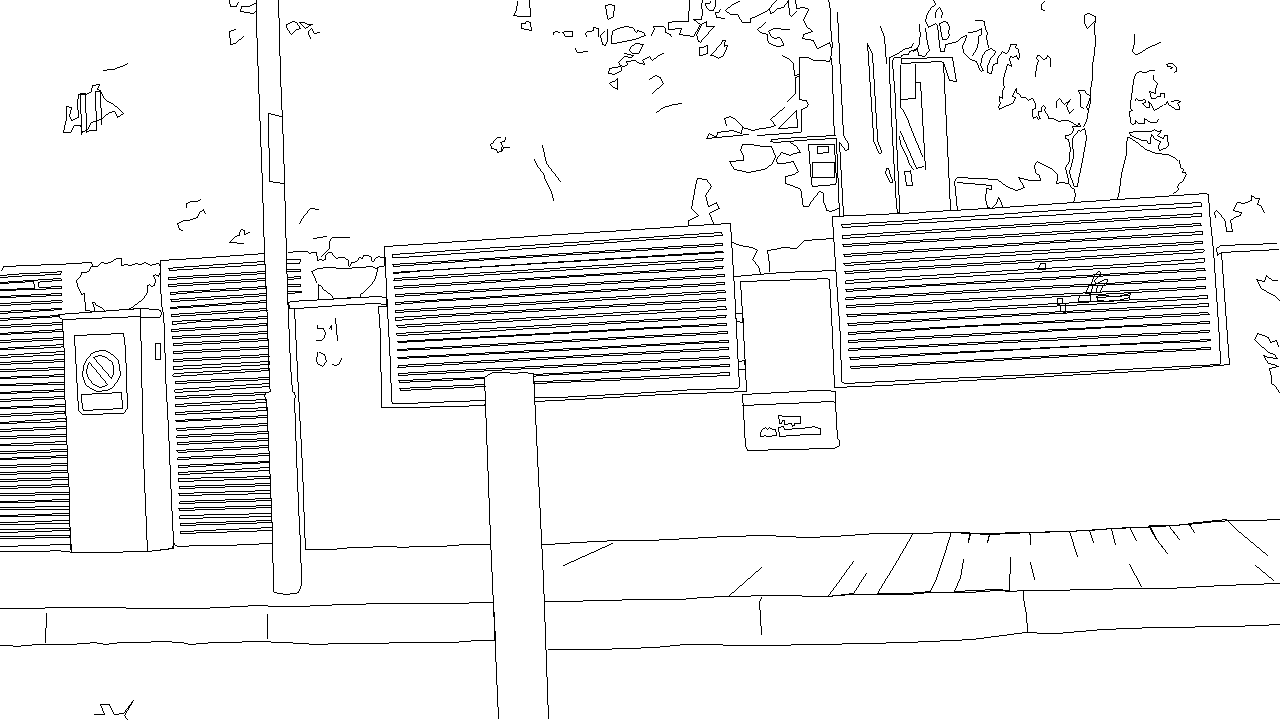}\\
			\vspace{0.05cm}
		\end{minipage}%
	}
	\subfigure[RCF~\cite{liu2019richer}]{
		\begin{minipage}[t]{0.22\textwidth}
			\centering
			\includegraphics[width=\textwidth]{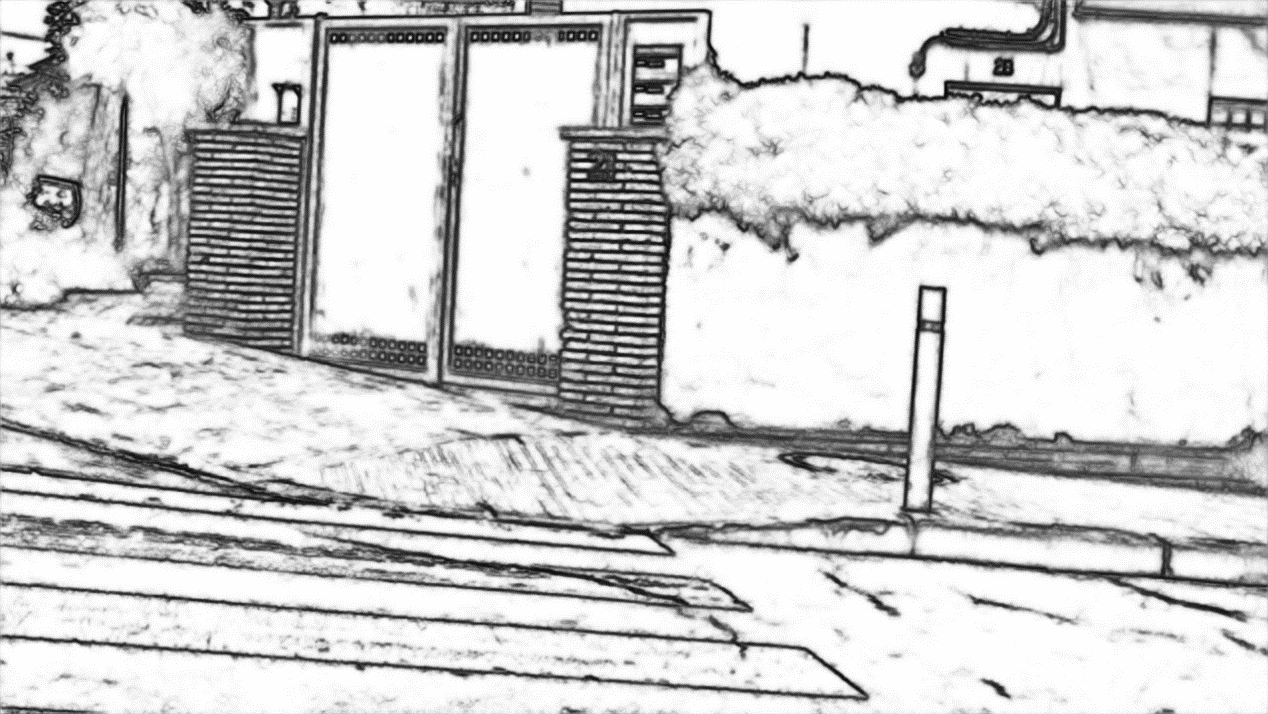}\\
			\vspace{0.05cm}
			\includegraphics[width=\textwidth]{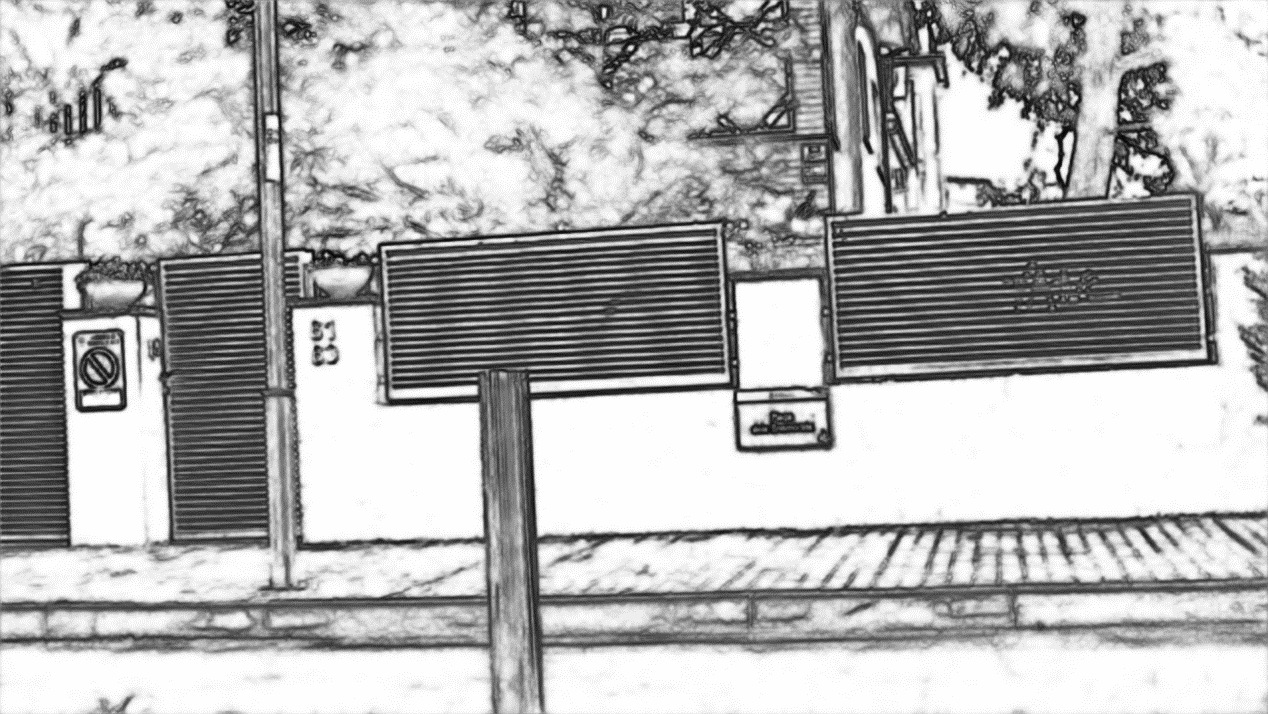}\\
			\vspace{0.05cm}
		\end{minipage}%
	}
	\subfigure[CTFN]{
		\begin{minipage}[t]{0.22\textwidth}
			\centering
			\includegraphics[width=\textwidth]{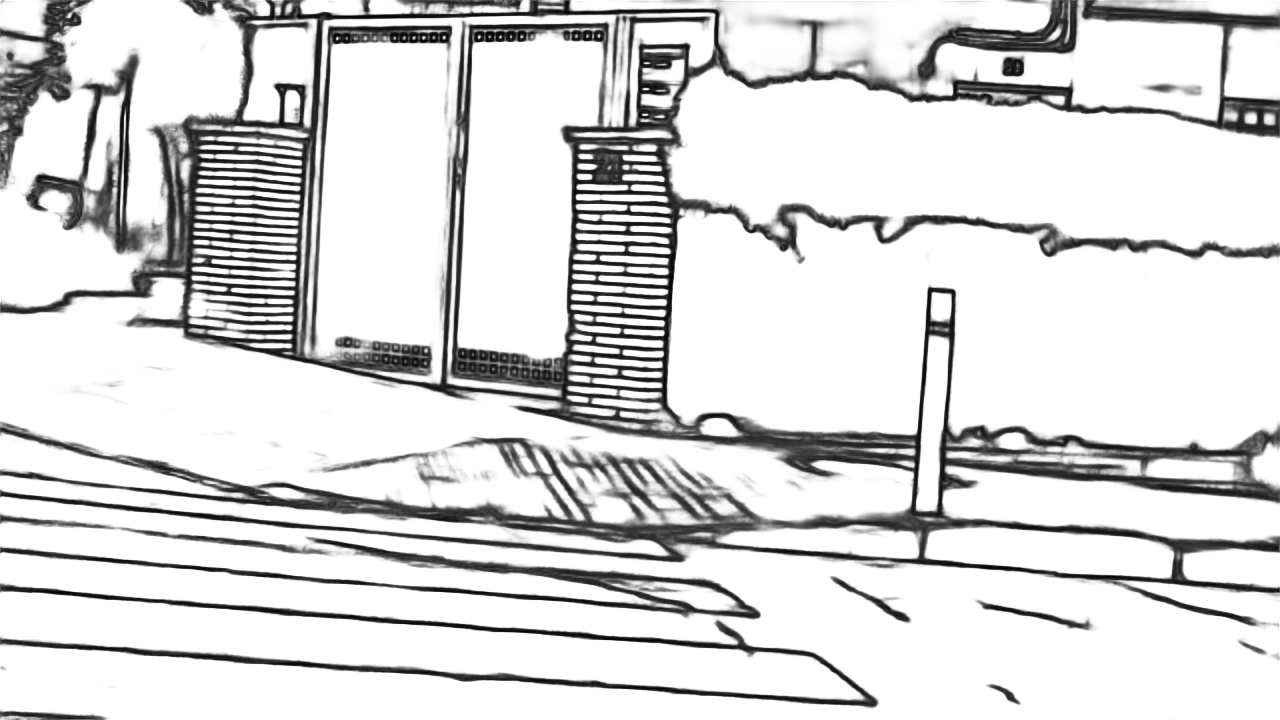}\\
			\vspace{0.05cm}
			\includegraphics[width=\textwidth]{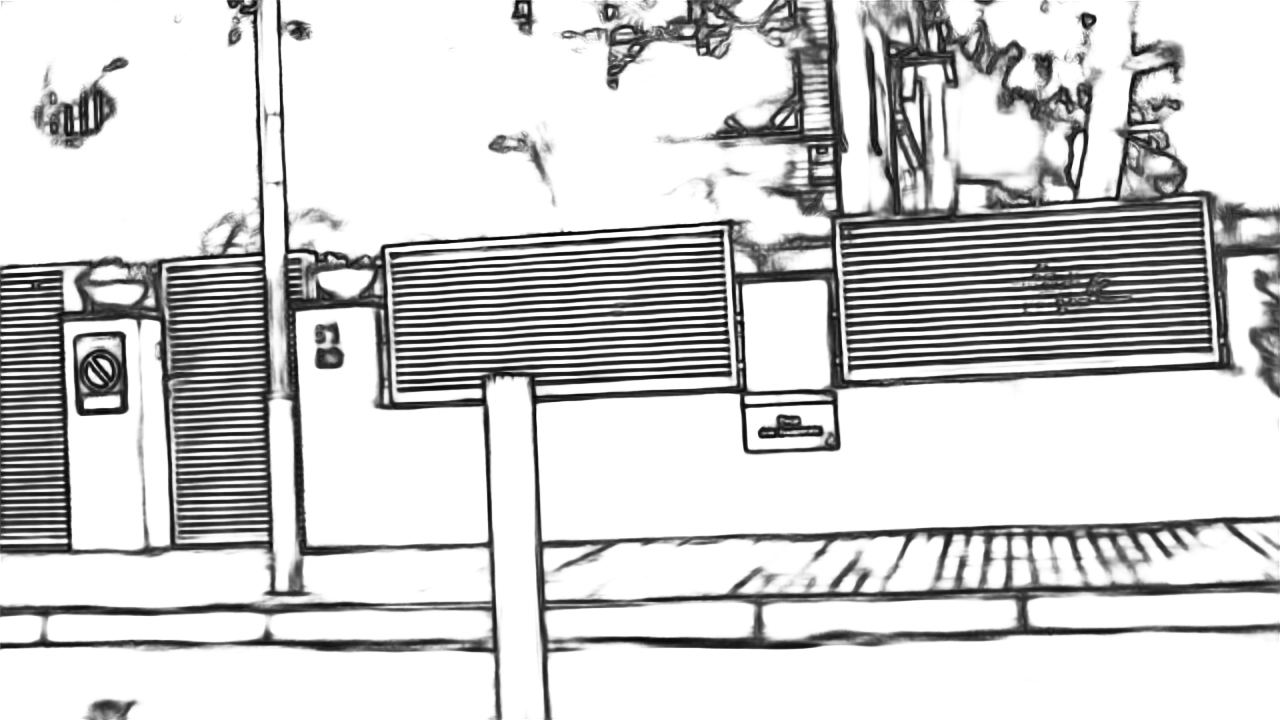}\\
			\vspace{0.05cm}
		\end{minipage}%
	}
	
	\caption{Comparison of edge detection results on BIPEDv2~\cite{soria2021bipedv2} dataset. All the results are raw edge maps before Non-Maximum Suppression.}\label{fig-biped}
	
\end{figure*}

We also record the evaluation results on BIPEDv2 and the comparison results with other methods are shown in Table~\ref{tab-biped}. Similarly, CTFN achieves promising results, with accuracy only 0.1\% lower than the best methods. We show some qualitative results in Fig.~\ref{fig-biped}. From left to right are the input images, ground truth, and the results of RCF and CTFN.

Model size will not change when tested on different datasets, and the FLOPs ratio of different models is also fixed. Therefore, we only show the accuracy of each method in NYUDv2 and BIPEDv2. As for model size and FLOPs, the results in Table~\ref{tab-bsds-vgg} can be referenced.


\subsection{Ablation Study}

The main innovations of CTFN are Semantic Enhancement Module (SEM) in the first feature fusion stage, Pixel-Level  Weighting (PPW) in the second feature fusion stage, and Dynamic Focal Loss (DFL). In this section, We verify the effectiveness of each module separately, as shown in Table~\ref{tab-abla}. The model is trained on BSDS500 train-val set and report the performance on test set.

\begin{table}[htbp]
	\begin{center}
    \caption{\centering{Ablation study on BSDS500 dataset.}}
    \label{tab-abla}
	\begin{tabular}{ccllll}
		\toprule
		Group 		& \multicolumn{3}{c}{Module} 	& ODS & OIS\\
		\midrule
		\multirow{3}*{1}     & SEM & Default  & DFL & 0.814 & 0.830\\
		& SEM & CoFusion & DFL & 0.817 & 0.832\\
        & SEM & CoFusion-l & DFL & 0.815 & 0.831\\
		& SEM & PPW      & DFL & 0.817 & 0.832\\
		\midrule
		\multirow{4}*{2} 	& SEM & PPW   & WCE 	& 0.810 & 0.828\\
		& SEM & PPW   & FL 		& 0.811 & 0.825\\
		& SEM & PPW   & FL-pre 	& 0.815 & 0.831\\
		& SEM & PPW   & DFL 	& 0.817 & 0.832\\
		\midrule
		\multirow{2}*{3} 	& -{}-& PPW  & DFL & 0.812 & 0.830\\
		& SEM & PPW  & DFL & 0.817 & 0.832\\
		\bottomrule
	\end{tabular}
	\end{center}
\end{table}

We first explore the impact of PPW. Experimental results are summarized in Group 1 of Table~\ref{tab-abla}. \textit{Default} represents the most commonly used weighted summation in previous edge detection methods~\cite{liu2019richer}. \textit{CoFusion} means the Context-aware fusion block proposed in the CATS~\cite{huan2021unmixing}. \textit{CoFusion-l} means \textit{CoFusion} with same number of channels as  \textit{PPW}. Compared with the weighted summation, both \textit{CoFusion} and \textit{PPW} lead to 0.3\% higher ODS. While \textit{PPW} takes 1/4 of \textit{CoFusion}'s parameters because of fewer channels.  Since the task is divided into channels and spaces in \textit{PPW}, each subtask is simpler and requires fewer parameters. When using same number of channels, \textit{PPW} outperforms \textit{Cofusion-l}. By comparison, The default Weighted summation uses channel weighting, \textit{Cofusion} uses mixed channel and space weighting, and \textit{PPW} uses separate channel and space weighting. The experimental results show that \textit{PPW} is the best choice.

As shown in Group 2 of Table~\ref{tab-abla}, we compare the impact of different loss functions. \textit{WCE} is the abbreviation of Weighted Cross-Entropy loss and \textit{FL} is the abbreviation of Focal Loss. \textit{FL-pre} means using WCE in the first epoch and then using FL to avoid confusion in the early stages of training, which is widely used. \textit{DFL} is the abbreviation of Dynamic Focal Loss. We can observe that the accuracy of DFL is significantly higher than WCE and FL. And the performance of FL-pre is slightly worse than DFL. In fact, FL-pre is a special case of DFL, in which the hyper-parameter $\mu\to0^+$. We can simulate the case by setting $\mu$ to $0.1^{-9}$ in Eq.~\ref{eq-omega'}.

\begin{table}[htbp]
    \begin{center}
    \caption{\centering{The effect of hyper-parameters $\gamma$ and $\mu$ on BSDS500 dataset.}}
     \label{tab-mu}
    \begin{tabular}{llll}
        \toprule
        $\gamma$ & $\mu$		& ODS 	&    OIS\\
        \midrule
        0		  & -{}-	    & 0.810 &	0.828\\
        \midrule
        1 		  & 0           & 0.814 &	0.829\\
        1         &$0.1^{-9}$   & 0.815 &   0.830\\
        1         & 0.5         & \textbf{0.817} &	\textbf{0.832}\\
        1         & 1           & 0.815 &   0.831\\
        \midrule
        2         & 0           & 0.813 &   0.829\\
        2         &$0.1^{-9}$   & 0.815 &   0.831\\
        2         & 0.5         & 0.816 &   0.831\\
        2 	      & 1           & 0.812 &   0.827\\
        \bottomrule
    \end{tabular}
    \end{center}
\end{table}
We further test the impact of hyper-parameters $\gamma$ and $\mu$. The results are summarized in Table.~\ref{tab-mu}. In Eq.~\ref{eq-omega'}, when $\gamma=0$, $\omega'$ is always equal to 1 irrespective of $\mu$, thus DFL degenerates into Weighted Cross-Entropy. Through a series of simple trials, it can be observed that as the values of $\gamma$ and $\mu$ increase, the accuracy increases first and then decreases. This evaluation is conducted to verify the effectiveness of DFL, thus more exhaustive explorations are not been done, even though they might lead to improvements in accuracy.

\begin{figure}[htbp]
	\centering
	\subfigure[stage 1]{
		\begin{minipage}[t]{0.2\linewidth}
			\centering
			\includegraphics[width=0.7\linewidth]{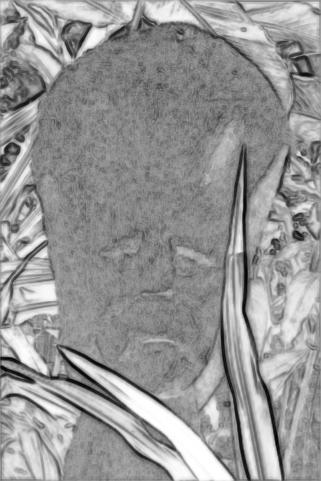}\\
			\vspace{0.05cm}
			\includegraphics[width=0.7\linewidth]{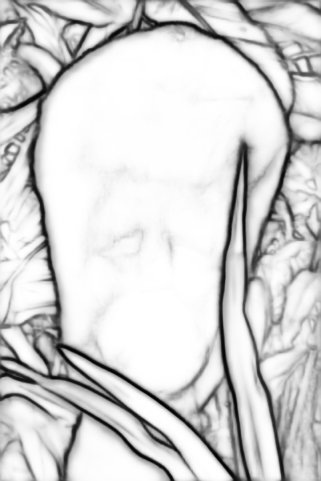}\\
			\vspace{0.05cm}
		\end{minipage}%
	}
	\hspace{-1.0cm}
	\subfigure[stage 2]{
		\begin{minipage}[t]{0.2\linewidth}
			\centering
			\includegraphics[width=0.7\linewidth]{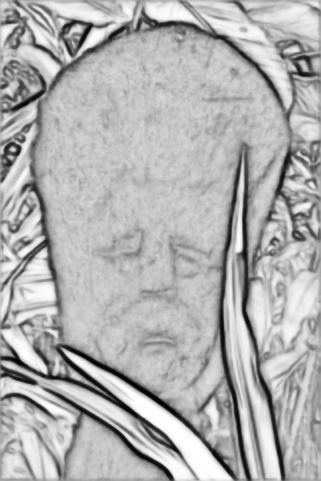}\\
			\vspace{0.05cm}
			\includegraphics[width=0.7\linewidth]{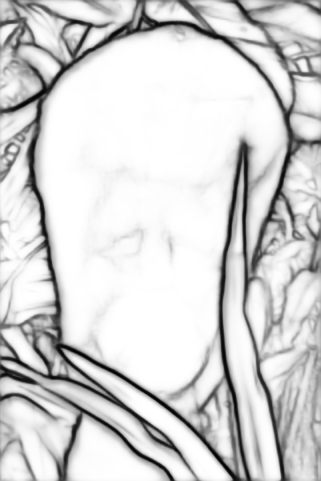}\\
			\vspace{0.05cm}
		\end{minipage}%
	}
	\hspace{-1.0cm}
	\subfigure[stage3]{
		\begin{minipage}[t]{0.2\linewidth}
			\centering
			\includegraphics[width=0.7\linewidth]{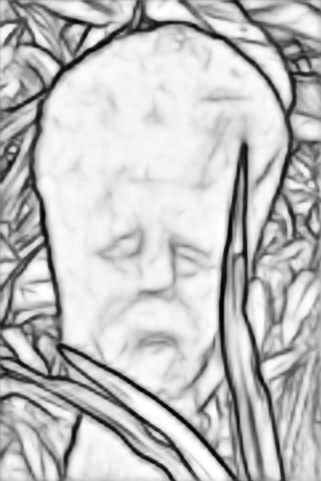}\\
			\vspace{0.05cm}
			\includegraphics[width=0.7\linewidth]{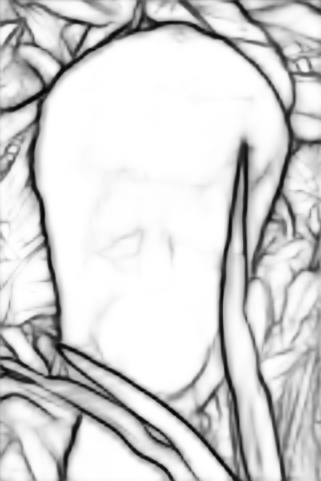}\\
			\vspace{0.05cm}
		\end{minipage}%
	}
	\hspace{-1.0cm}
	\subfigure[stage 4]{
		\begin{minipage}[t]{0.2\linewidth}
			\centering
			\includegraphics[width=0.7\linewidth]{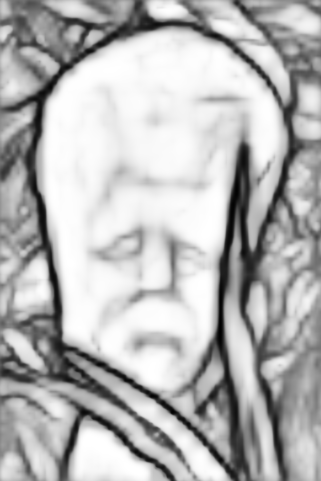}\\
			\vspace{0.05cm}
			\includegraphics[width=0.7\linewidth]{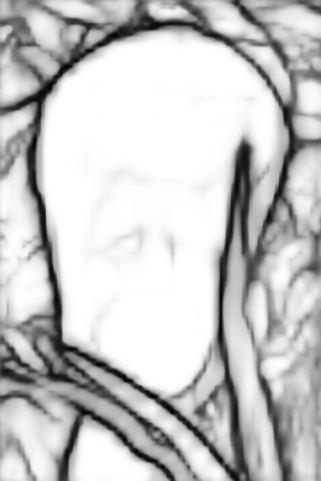}\\
			\vspace{0.05cm}
		\end{minipage}%
	}
	\hspace{-1.0cm}
	\subfigure[stage 5]{
		\begin{minipage}[t]{0.2\linewidth}
			\centering
			\includegraphics[width=0.7\linewidth]{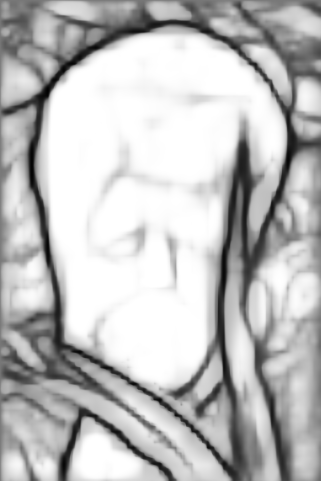}\\
			\vspace{0.05cm}
			\includegraphics[width=0.7\linewidth]{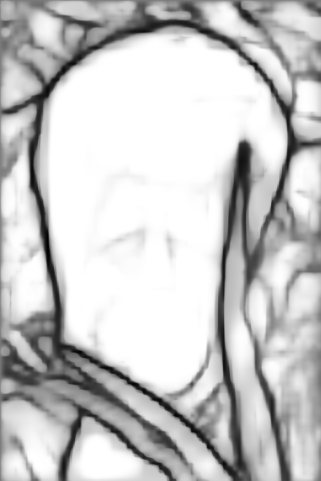}\\
			\vspace{0.05cm}
		\end{minipage}%
  }
  \hspace{-1.0cm}
	\subfigure[fusion]{
		\begin{minipage}[t]{0.2\linewidth}
			\centering
			\includegraphics[width=0.7\linewidth]{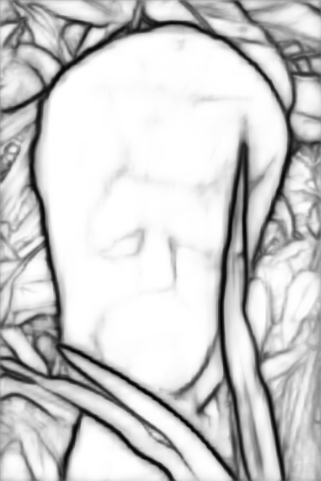}\\
			\vspace{0.05cm}
			\includegraphics[width=0.7\linewidth]{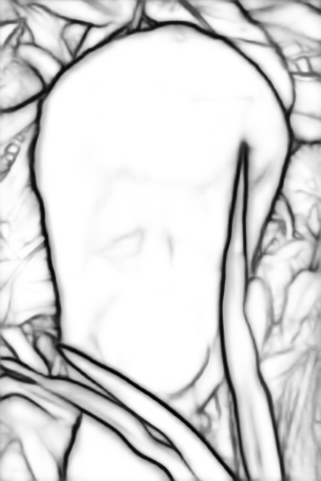}\\
			\vspace{0.05cm}
		\end{minipage}%
	}
	
	\caption{Comparison of different stage results. The first row is the results of the model not including SEM, and the second row is the results of the model including SEM. All the results are raw edge maps before Non-Maximum Suppression. Stages 1 to 5 correspond to the output of the model at different scales, and \emph{fusion} is the final output of the model, the fusion of stages 1 to 5.}
 \label{fig-cmp-sem}
	
\end{figure}

We verify the effectiveness of SEM in group 3 of Table~\ref{tab-abla}. It can be observed that SEM improves the ODS of the model by 0.5\%, which can effectively improve the model performance. The visualization results in Fig.~\ref{fig-cmp-sem} further validate our conclusion. The effect of SEM is more notorious for fine-scale branches, where there are more textures when SEM is not used.

\subsection{Visual Analysis}

\begin{figure}[htbp]
	\begin{center}
	\subfigure[image]{
		\includegraphics[width=0.3\linewidth]{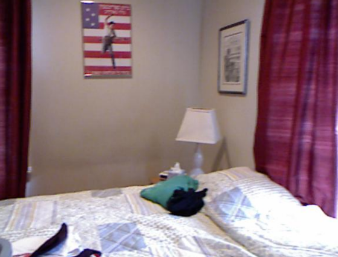}}
	\subfigure[o$_1$]{
		\includegraphics[width=0.3\linewidth]{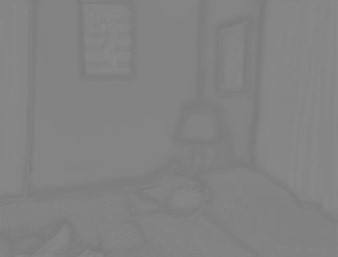}}
	\subfigure[$\omega_1$]{
		\includegraphics[width=0.3\linewidth]{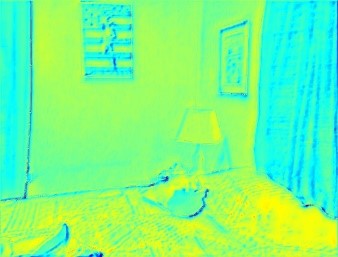}
        \label{fig-DFLvis-omg1}}
    \subfigure[ground truth]{
    	\includegraphics[width=0.3\linewidth]{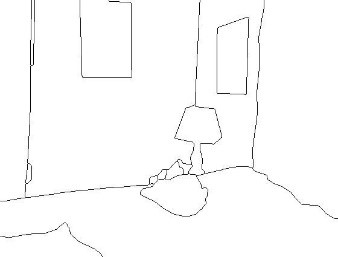}
        \label{fig-DFLvis-gt}}
	\subfigure[o$_2$]{
		\includegraphics[width=0.3\linewidth]{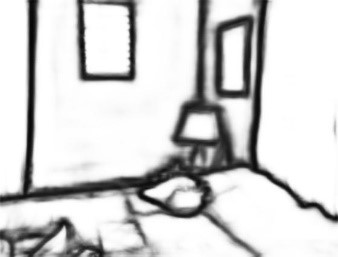}}
	\subfigure[$\omega_2$]{
        \label{fig-DFLvis-omg2}
		\includegraphics[width=0.3\linewidth]{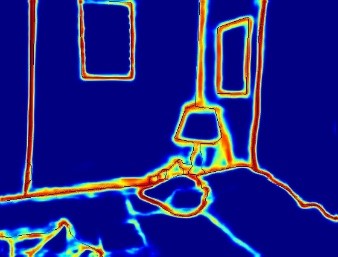}}
  \end{center}
	\caption{Visualization of Focal Loss (FL). (a) An image from the NYUDv2 dataset, (b) The output of CTFN in the early training, (c) The weight of FL in the early training stage, (d) Ground truth of the image, (e) The output of CTFN in the stable training, (f) the weight of FL in the stable training stage} 
	\label{fig-DFLvis}
	
\end{figure}
We visualize the weight $\omega$ of Focal Loss in Fig.~\ref{fig-DFLvis}, which is mentioned in Eq.~\ref{eq-fl}. We can observe that in the early training stage, the output of the model is disordered, leading to similar and chaotic weight $\omega_1$. Hard samples are not paid enough attention to. While the situation is quite different in the stable training stage, the misclassified negative samples near the edge and the error-detected texture in the background correspond to greater weight, which contributes to better edge map generation. By comparing Fig.~\ref{fig-DFLvis-omg1} and Fig.~\ref{fig-DFLvis-omg2}, it can be seen that the error of $\omega$ in the early training stage is relatively larger, so the confidence margin of $\omega$ should be low and gradually increase with the training of the model, which is also the principle of Dynamic Focal Loss.

Another point of note is that the transition from the edge to the non-edge area is softer in the results of CTFN. Due to the imbalance of positive and negative sample weights in Weighted Cross-Entropy Loss, the existing methods suffer from blurry edges~\cite{deng2020deep,huan2021unmixing}. As shown in Fig.~\ref{fig-bsds}, the outputs of these methods are far from single-pixel edges. Even though the edges are processed to a single-pixel width in subsequent Non-Maximum Suppression (NMS), blurry edges lead to localization ambiguity and is detrimental to the final accuracy. While DFL is utilized in CTFN to assign larger weights to the non-edge regions near the edges due to the large difference between the non-edges and the ground truth, which is shown in Fig.~\ref{fig-DFLvis-omg2}. 

\begin{figure}[htbp]
\centering
\subfigure[$left$: Ground truth, $middle$: The result of CTFN; $right$: The result of BDCN.]{
\fboxsep=0pt
\fbox{\includegraphics[width=0.3\linewidth]{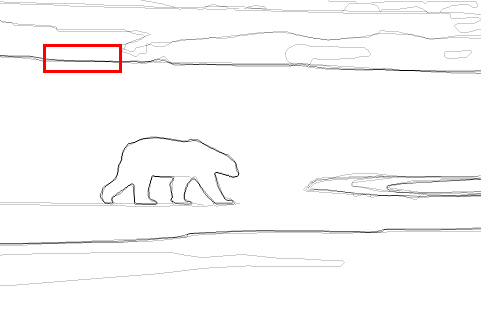}}
\fbox{\includegraphics[width=0.3\linewidth]{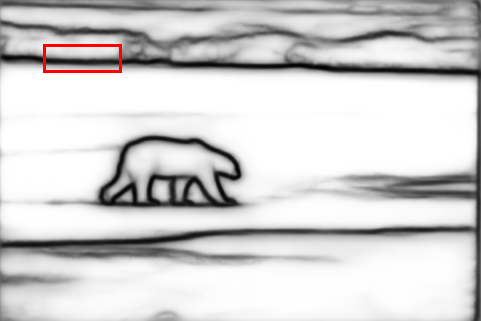}}
\fbox{\includegraphics[width=0.3\linewidth]{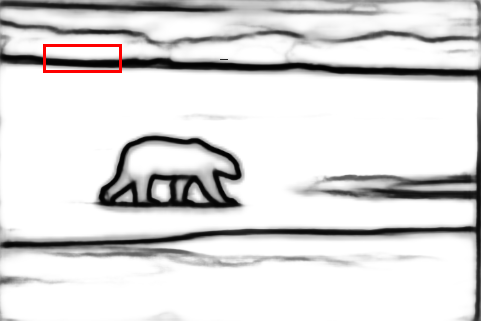}}
\label{fig-DFLWCE-a}
}
\quad
\subfigure[Display of details in red boxes of Fig.(a). $left$: Ground truth, $middle$: The result of CTFN; $right$: The result of BDCN.]{
\fboxsep=0pt
\fbox{\includegraphics[width=0.3\linewidth]{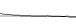}}
\fbox{\includegraphics[width=0.3\linewidth]{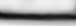}}
\fbox{\includegraphics[width=0.3\linewidth]{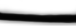}}
\label{fig-DFLWCE-b}
}
\quad
\subfigure[The images in Fig.(b) is treated with non-maximum suppression (NMS), and the difference between the results of CTFN and BDCN can be more obviously observed. $left$: Ground truth, $middle$: The result of CTFN; $right$: The result of BDCN.]{
\fboxsep=0pt
\fbox{\includegraphics[width=0.3\linewidth]{Fig-cmp_softer-GT_s.png}}
\fbox{\includegraphics[width=0.3\linewidth]{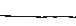}}
\fbox{\includegraphics[width=0.3\linewidth]{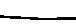}}
\label{fig-DFLWCE-c}
}
\quad
\subfigure[The result of concatenating the images in Fig.(c) in the channel dimension. The image is binarized for a better observation. Purple denotes the result of the CTFN and cyan denotes the groundtruth, yellow indicates the result of BDCN, and the other colors are derived from the mixture of these three colors.]{
\fboxsep=0pt
\fbox{\includegraphics[width=0.9\linewidth]{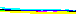}}
\label{fig-DFLWCE-d}
}
\caption{Comparison of differences in visual effects between CTFN and BDCN.}
\label{fig-DFLWCE}
\end{figure}

We visualize the results of CTFN and a typical WCE-based method BDCN~\cite{he2022bi-directional} in Fig.~\ref{fig-DFLWCE-a}. Compared with the ground truth on the left, the results of the two models seem to differ little. However, when we show the details of them in Fig.~\ref{fig-DFLWCE-b}, we can observe that the edge of BDCN is thicker, which leads to larger localization ambiguity after NMS, as shown in Fig.~\ref{fig-DFLWCE-c}. This contrast can be observed more clearly after concatenating them in the channel dimension, as shown in Fig.~\ref{fig-DFLWCE-d}. We can observe that the result of the CTFN (purple) deviate less from the ground truth (cyan) than the result of BDCN (yellow).
Therefore, we can conclude that CTFN can effectively alleviate the problem of edge localization ambiguity.

\section{Conclusion}
In this paper, we review existing deep-learning based edge detection methods and propose a new Compact Twice Fusion Network(CTFN), in which we divide the edge detection model into three parts: the backbone, the first feature fusion stage, and the second feature fusion stage. We propose two lightweight modules SEM and PPW to fuse multi-scale features and further introduce a dynamic focal loss to focus on the hard samples of images. Experimental results on multiple datasets verify the effectiveness of CTFN. Compared to state-of-the-art methods, CTFN achieves competitive accuracy and higher efficiency.

\textbf{Limitation.} For fair comparison, our method still uses VGG-16 as the backbone, which accounts for more than 95\% of the model parameters. This limits the further compression of the model. And the feature extraction ability of VGG-16 is hard to meet the needs of edge detection. In addition, compared with WCE, DFL adds two hyperparameters, which should be different on different datasets. This makes it more difficult for our method to transfer on different datasets. Therefore, in the future work, we will explore more efficient backbone and design a dynamic Focal Loss with adaptive hyperparameters.

\backmatter

\bmhead*{Acknowledgments}
This work is partly supported by National key r\&d program(Grant no. 2019YFF0301800),National Natural Science Foundation of China (Grant no. 61379106),the Shandong Provincial Natural Science Foundation (Grant nos.ZR2013FM036,ZR2015FM011). Xavier Soria was funded by the Air Force Office of Scientific Research under Award FA9550-22-1-0261.

\section*{Declarations}

\begin{itemize}


\item Competing interests

No potential conflicts of interest were identified.







\item Availability of data and materials

This work used publicly available dataset for the training and validation. The code source is available.



\end{itemize}

\bibliography{sn-bibliography}


\end{document}